%% file: main.tex
\documentclass[runningheads]{llncs}
\usepackage[T1]{fontenc}
\usepackage{xspace}
\def\ourName{SimEdit\xspace}
\usepackage{graphicx}
\usepackage{booktabs}
\usepackage[misc]{ifsym}

\usepackage{cite}
\usepackage{amsmath}
\usepackage{amssymb}
\usepackage{multirow}
\usepackage{multicol}
\usepackage{algorithm}
\usepackage{algorithmic}
\usepackage[most]{tcolorbox}
\tcbuselibrary{skins,breakable}
\usepackage{graphicx}
\usepackage{tikz}
\usepackage{xcolor}
\definecolor{Red}{RGB}{255, 199, 206}
\definecolor{Blue}{RGB}{221, 235, 247}
\definecolor{Yellow}{RGB}{255, 235, 156}
\definecolor{softgreen}{RGB}{34,139,34}
\usepackage[hidelinks]{hyperref}

\makeatletter
\def\@fnsymbol#1{\ensuremath{\ifcase#1\or \dagger\or \ddagger\or
   \mathsection\or \mathparagraph\or \|\or **\or \dagger\dagger
   \or \ddagger\ddagger \else\@ctrerr\fi}}
\makeatother

\begin{document}


\title{Conditioning Matters: Stabilizing Inversion and Attention in Diffusion Image Editing}

\titlerunning{Conditioning Matters for Diffusion Image Editing}

\author{
Zheyuan Zhan$^{1,2,3}$, Hongchen Li$^{3}$, Can Wang$^{1,2,3}$, Yinfei Ma$^{3}$\\ Mingzhen Huang$^{4}$, Ruoshi Bai$^{3}$, Jiawei Chen$^{1,2,3}$, Siwei Lyu$^{4}$, Defang Chen$^{4}$\thanks{Corresponding author} \\
}
\tocauthor{Zheyuan Zhan, Hongchen Li, Can Wang, Yinfei Ma, Mingzhen Huang, Ruoshi Bai, Jiawei Chen, Siwei Lyu, Defang Chen}
\toctitle{Conditioning Matters: Stabilizing Inversion and Attention in Diffusion Image Editing}


\authorrunning{Z. Zhan et al.}

\institute{$^1$State Key Laboratory of Blockchain and Data Security, Zhejiang University\\
$^2$HangZhou High-Tech Zong (Binjiang) Institute of Blockchain and Data Security\\
$^3$College of Computer Science, Zhejiang University\\
$^4$University at Buffalo, State University of New York
}

\maketitle              

\input{sec/0_abstract}
\begin{figure*}[!h]
\centering
\hspace{-0.5in}
\begin{tikzpicture}
    
  \node[anchor=south west,inner sep=0] (img) at (0,0)
    {\includegraphics[width=1\linewidth]{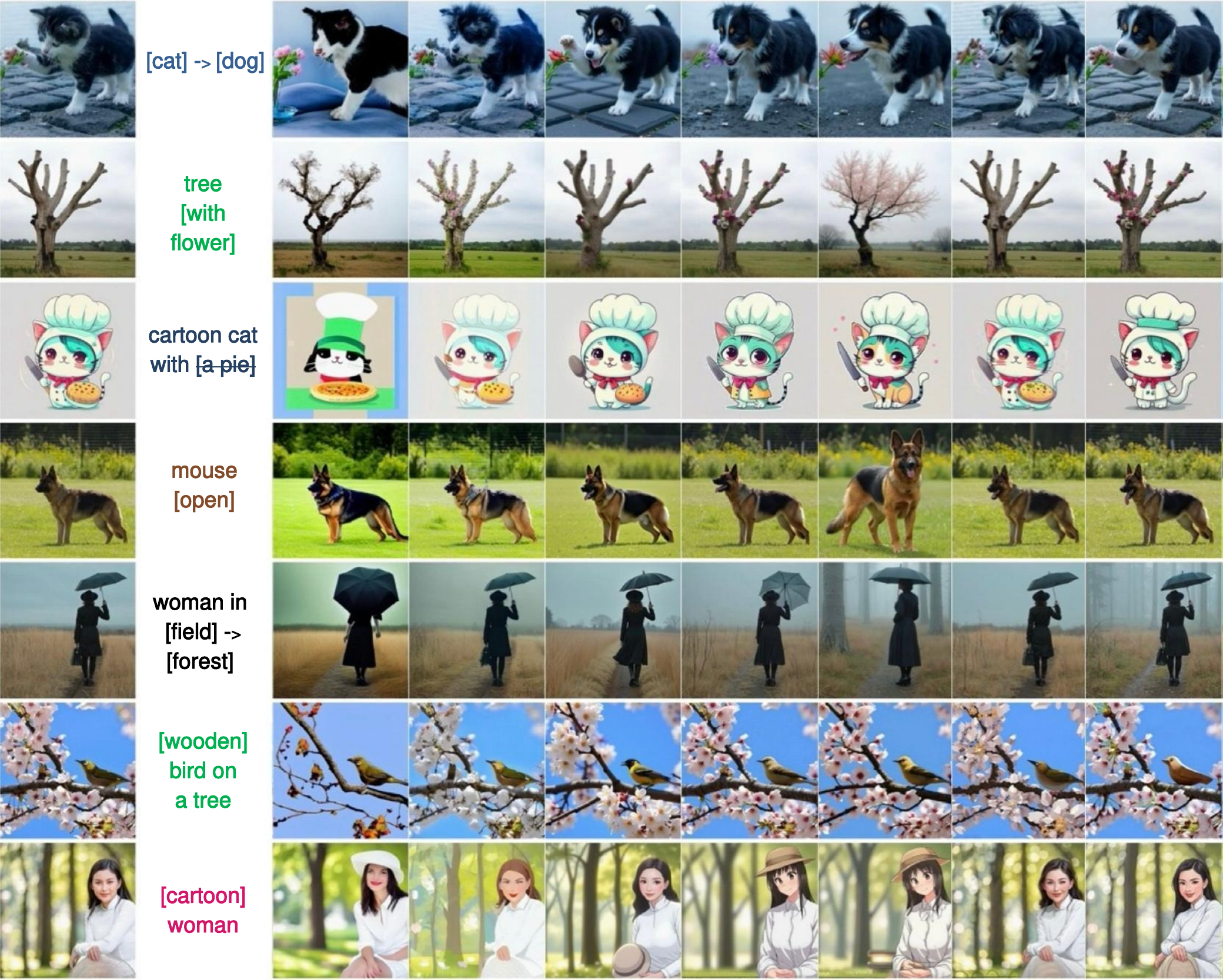}};

  \begin{scope}[x={(img.south east)},y={(img.north west)}]

    \def\leftstrip{0} 
    \def\ncols{9}             
    \def\ytext{1.01}          

    \foreach \i/\t in {
      1/Source,
      2/  ,
      3/P2P, 
      4/PnP, 
      5/RF-Inversion, 
      6/RF-Edit, 
      7/FireFlow, 
      8/UniEdit, 
      9/Ours,
    }{
      \pgfmathsetmacro\x{\leftstrip + (\i-0.5)*(1-\leftstrip)/\ncols}
      \node[font=\tiny\bfseries,align=center,
      ] at (\x,\ytext) {\t};
    }
  \end{scope}
\end{tikzpicture}
\caption{We compare our proposed \ourName with previous attention manipulation-based editing methods~\cite{hertz2023p2p,tumanyan2023plug,rout2025semantic,wang2025taming,deng2025fireflow,jiao2025uniedit} across seven scenarios. 
Overall, our method achieves better performance in both background preservation and semantic fidelity. 
}
\label{fig:editing}
\end{figure*}

\input{sec/1_introduction}

\input{sec/2_related_works}
\input{sec/3_preliminaries}

\input{sec/4_new_motivation}

\input{sec/4_method}

\input{sec/5_experiments}
\input{sec/6_conclusion}

\begin{credits}
\subsubsection{\ackname}
This work is supported by the National Natural Science Foundation of China (No.~62476244), the Starry Night Science Fund of Zhejiang University Shanghai Institute for Advanced Study, China (Grant No.~SN-ZJU-SIAS-001) and the advanced computing resources provided by the Supercomputing Center of Hangzhou City University. We also thank Jialing Cai for the help with figure preparation.
\end{credits}

{
    \small
    \bibliographystyle{splncs04}
    \bibliography{main}
}
\input{appendix} 
\end{document}

%% file: sec/0_abstract.tex
\begin{abstract}

Inversion-based image editing offers flexible and training-free control but still struggles with inversion accuracy and the trade-off between editing fidelity and background preservation. While recent methods improve inversion formulations or attention interactions, the role of textual conditioning in shaping diffusion dynamics and editing behavior remains underexplored. We show both empirically and theoretically that the precision of textual conditioning influences inversion stability by modulating the geometry of the diffusion velocity field, while also affecting the consistency of cross-branch attention during editing. These effects directly impact background preservation and semantic fidelity. Building on this analysis, we propose SimEdit, a conditioning-aware framework with two complementary components: (a) \textit{conditioning refinement}, which constructs conditioning signals with improved semantic precision and structural alignment to facilitate stable inversion and consistent attention manipulation, and (b) \textit{token-wise cross-branch attention control}, which separates edit-relevant and structure-preserving components and modulates them asymmetrically during attention manipulation. Extensive experiments on PIE-Bench demonstrate that SimEdit consistently improves both inversion reconstruction quality and editing performance over previous attention-manipulation approaches. Our code is available at \url{https://github.com/zju-pi/SimEdit}.


\end{abstract}

%% file: sec/1_introduction.tex
\section{Introduction}\label{sec:introduction}

Text-guided image editing~\cite{kawar2023imagic,hertz2023p2p} modifies a given source image based on textual instructions, aiming for precise semantic changes while preserving editing-irrelevant content. The source image can be real or synthesized, making this task widely applicable. Most state-of-the-art image editing methods are built upon diffusion models due to their excellent controllability and sampling diversity~\cite{rombach2022ldm,esser2024scaling,black-forest2024flux,zhan2025conditional}. However, retraining or fine-tuning diffusion models for task-specific editing remains computationally expensive.

Training-free approaches based on inversion and attention manipulation~\cite{hertz2023p2p,tumanyan2023plug,cao2023masactrl,patashnik2023localizing,lu2023tf} provide a flexible alternative, yet they face two fundamental challenges: (1) inversion errors accumulate~\cite{mokady2023null,ju2023direct,xu2025fteedit}, degrading reconstruction accuracy and attention fidelity; and (2) cross-branch attention features often diverge significantly in semantics or spatial layout, leading to poor background preservation and structural distortions~\cite{hertz2023p2p,wang2025taming}. 
To address these challenges, existing methods mainly focus on improving inversion solvers or attention manipulation mechanisms, while largely overlooking the role of textual conditioning during the editing process. 
In fact, we observe that textual conditioning shapes the diffusion velocity field and attention behavior, thereby affecting both inversion trajectories and downstream attention manipulation.
In this work, we systematically analyze how conditioning precision affects the stability of diffusion dynamics. We demonstrate both empirically and theoretically that coarse textual conditioning leads to an unstable diffusion velocity field with higher variance and larger effective Lipschitz constants, resulting in larger trajectory deviations and reconstruction errors. In contrast, more precise and semantically grounded conditioning stabilizes the velocity geometry. We further observe that conditioning precision and structural alignment also improve the consistency of cross-branch attention features during editing, which enhances the effectiveness of attention manipulation.

Building on these observations, we propose \ourName, a \underline{sim}ple structured conditioning framework for inversion-based diffusion \underline{edit}ing. Guided by the analysis on the correlation between textual conditioning and diffusion velocity field, we first refine conditioning signals to improve inversion stability and cross-branch structural consistency. However, refined conditioning also increases token density, which makes attention manipulation more delicate: editing requires simultaneously preserving structure-preserving content and enforcing edit-driving semantics, two fundamentally different roles that cannot be handled uniformly. We therefore decompose conditioning components into edit-relevant and edit-irrelevant groups, and modulate them differently in attention via token-wise cross-branch attention control to balance structural appearance preservation with precise semantic transfer.
Through this conditioning-aware perspective, \ourName achieves consistent improvements in both inversion accuracy and editing fidelity on PIE-Bench as demonstrated in Fig.\ref{fig:editing}. Moreover, our approach is training-free and can be seamlessly integrated into existing inversion-based pipelines, offering a practical yet theoretically grounded enhancement for diffusion-based image editing.
Our main contributions are summarized as follows:
\begin{itemize}

\item We present a geometric perspective on textual conditioning in inversion process, empirically and theoretically demonstrating that conditioning precision directly influences the smoothness of the diffusion velocity field and the stability of inversion trajectories.

\item Based on this insight, we propose \ourName, a conditioning-aware framework that refines conditioning signals and introduces an asymmetric attention modulation strategy for structure-preserving and edit-driving conditioning components to handle them differently, thereby balancing structural preservation and semantic editing.

\item Extensive experiments on PIE-Bench demonstrate that \ourName consistently improves inversion reconstruction quality and editing fidelity across multiple attention manipulation pipelines, while remaining fully training-free and compatible with existing methods.

\end{itemize}

%% file: sec/2_related_works.tex
\section{Related works}

Text-to-image diffusion models rely on textual conditioning to guide generative dynamics through cross-attention or transformer-based conditioning mechanisms~\cite{rombach2022ldm,peebles2023scalable,esser2024scaling,black-forest2024flux}. Recent DiT-based architectures achieve stronger synthesis quality and semantic coherence via deeper transformer backbones and more expressive conditioning pathways~\cite{esser2024scaling}. While prior studies analyze how attention maps correlate with semantic regions~\cite{tang2023daam} or how conditioning affects controllability~\cite{datta2024prompt}, the geometric influence of textual conditioning on diffusion dynamics remains underexplored. In particular, little attention has been paid to how the precision and structure of conditioning influence diffusion dynamics, inversion stability and downstream editing behavior.

Diffusion-based inversion enables training-free image editing by projecting a source image into latent space and regenerating it under modified conditions~\cite{su2023ddib,meng2022SDEdit}. However, inversion fidelity critically depends on the stability of diffusion trajectories. Accumulated discretization errors and integration mismatch often lead to divergence between inversion and reconstruction paths~\cite{mokady2023null,ju2023direct,wu2023latent,nie2023blessing}. To mitigate these issues, previous works explore improved solvers~\cite{lu2023tf,brack2023ledits++,wang2025taming}, trajectory alignment strategies~\cite{mokady2023null,ju2023direct}, and stochastic inversion techniques~\cite{wu2023latent,nie2023blessing}. These approaches primarily focus on improving the inversion process itself, while largely treating the conditioning signal as fixed.

Attention manipulation further enhances editing controllability by injecting or reweighting features between source and target branches. Prompt-to-Prompt (P2P)~\cite{hertz2023p2p} transfers cross-attention maps to preserve spatial structure, while subsequent methods extend this idea using self-attention maps~\cite{tumanyan2023plug}, value matrices~\cite{cao2023masactrl}, segmentation masks~\cite{patashnik2023localizing,lu2023tf}, or stronger base models~\cite{rout2025semantic,wang2025taming}. Although these methods improve background preservation and semantic control, they rely on stable inversion trajectories and consistent cross-branch attention features to function effectively.

In contrast to prior works that refine inversion solvers or modulate attention features, we investigate the geometric role of textual conditioning in stabilizing diffusion dynamics and demonstrate that conditioning-aware representations of textual conditioning signals can substantially improve both inversion stability and attention consistency without modifying the underlying diffusion model.

%% file: sec/3_preliminaries.tex
\section{Background}

\textbf{Flow Matching}.
Recent state-of-the-art text-to-image synthesis and editing methods~\cite{esser2024scaling,black-forest2024flux,tewel2024add,wang2025taming} increasingly adopt \textit{flow matching} to model the generative process. In this formulation, flow matching learns a continuous mapping from a data distribution $\mathbf{x}_0 \sim p_0$ to a prior $\mathbf{x}_1 \sim p_1$~\cite{lipman2023flow}. A linear interpolation $\mathbf{x}_{\tau} = (1 - \tau)\mathbf{x}_0 + \tau \mathbf{x}_1$, with $\tau \in [0, 1]$, defines the forward trajectory, and the corresponding velocity field satisfies $\frac{d \mathbf{x}_\tau}{d\tau} = \mathbf{v}_\theta(\mathbf{x}_\tau, \tau \mid P)$. The velocity field $\mathbf{v}_\theta(\mathbf{x}_\tau, \tau \mid P)$ defines the time derivative of the sampling trajectory conditioned on textual signal $P$, which governs how the latent $\mathbf{x}_{\tau}$ evolves at timestep $\tau$.
Flow matching trains a time-dependent conditioned velocity field $\mathbf{v}_{\theta}(\mathbf{x}_\tau, \tau \mid P)$ via the objective: $
\mathcal{L}_{\theta} = \mathbb{E}_{\tau,\ \mathbf{x}_0,\ \mathbf{x}_1,\ P} 
\left[ \| (\mathbf{x}_1 - \mathbf{x}_0) - \mathbf{v}_\theta(\mathbf{x}_\tau, \tau \mid P) \|^2 \right].$
The prior $p_1$ is typically set to the standard Gaussian $\mathcal{N}(\mathbf{0}, \mathbf{I})$, aligning with diffusion models~\cite{lipman2023flow}\footnote{In this paper, we interchangeably use (Gaussian) flow-based and diffusion-based generative models due to their equivalence.}. In practice, the continuous trajectory is discretized into a predefined schedule $\{\sigma_T, \ldots, \sigma_0\}$, yielding the sampling update: $
\mathbf{x}_{\sigma_t}=\mathbf{x}_{\sigma_{t+1}}+(\sigma_t-\sigma_{t+1}) \cdot \mathbf{v}_\theta(\mathbf{x}_{\sigma_{t+1}}, \sigma_{t+1} \mid P)$. Importantly, the conditioning signal $P$ modulates the learned velocity field and thus shapes the geometry and stability of diffusion trajectories. The inversion trajectory $\mathbf{x}_{\sigma_t}^{\text{inv}}$ in the flow matching framework first projects the clean source image to a noise latent conditioned on the source prompt $P_{\text{src}}$, and then a reconstruction process is performed conditioned on the same prompt to collect attention features for injection. However, due to accumulated discretization errors, the reconstruction trajectory gradually deviates from the inversion trajectory, leading to inferior recovery of the source image in attention manipulation-based editing.

\textbf{Inversion-based editing}.
Recent T2I models \cite{esser2024scaling,black-forest2024flux} integrate text–image interactions and intra-image dependencies within unified attention blocks that process both textual and visual tokens. 
Starting from the inverted latent noise $\mathbf{x}_{\sigma_T}^{\text{inv}}$, we can perform sampling conditioned on the target prompt $P_{\text{tar}}$ to generate an image that loosely preserves the structure of the source image. However, this direct approach often fails to maintain fine-grained details of $I_{\text{src}}$ due to the limited control during the sampling process. To address this, attention manipulation in inversion-based editing methods operates on two parallel sampling trajectories starting from the same inverted latent: the \textbf{source (reconstruction) branch} conditioned on $P_{\text{src}}$ that reconstructs the original image, and the \textbf{target (editing) branch} conditioned on $P_{\text{tar}}$ that generates the edited result. These methods inject attention features—such as queries, keys, and values—extracted from the source branch into the target branch~\cite{hertz2023p2p,tumanyan2023plug,cao2023masactrl}. Specifically, these methods linearly interpolate attention features at selected layers and timesteps, with interpolation weights controlling the contribution from each branch. The resulting mixed attention is then used during the target branch's sampling, guiding generation with structural cues from the source image while adhering to the semantics of the target prompt. In our framework, we study multiple attention feature replacement strategies. Empirical comparisons show that our conditioning-aware structuring consistently improves inversion stability and editing performance across different replacement schemes.

%% file: sec/4_new_motivation.tex
\section{Motivation}\label{sec:motivation}

\begin{figure}[t]
    \centering
    \includegraphics[width=0.95\linewidth]{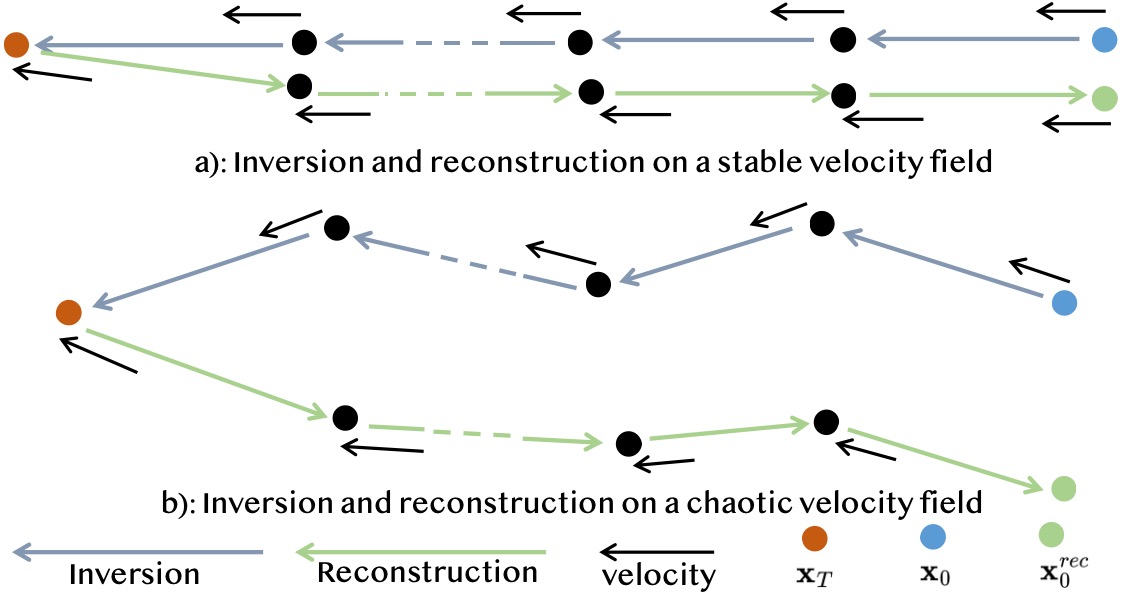}
    \caption{Illustration of inversion and reconstruction under (a) stable and (b) chaotic velocity fields. A more stable velocity field leads the forward and backward integration trajectories to exhibit closer alignment, resulting in smaller reconstruction errors.}
    \label{fig:inversion_sta}
\end{figure}

In practice, inversion-based image editing is a condition-driven process consisting of two coupled stages: \textit{inversion} and \textit{cross-branch attention manipulation}. 

In the inversion stage, the source image is mapped into the latent space by integrating a condition-dependent velocity field. The properties of the textual conditioning influence the stability of this velocity field, and a more stable velocity field leads to a more consistent inversion trajectory, which in turn determines how accurately the structural and semantic information can be reconstructed. In the subsequent attention manipulation stage, features extracted from the reconstruction branch are injected into a target-conditioned branch initialized from the same latent state obtained from inversion. The effectiveness of this cross-branch attention manipulation is influenced by the form of the textual conditioning signals, which affects how preserved structures and newly introduced semantics are combined during editing.

From this perspective, textual conditioning is not merely a descriptive input, but an important factor that shapes both the velocity field during inversion and the feature coupling during attention manipulation. In the following, we analyze how its precision and structure modulate (i) the stability of the inversion process and (ii) the effectiveness of cross-branch attention manipulation, providing a unified perspective on the role of textual conditioning in inversion-based editing.

\begin{figure*}[t]
    \centering
    \includegraphics[width=1\linewidth]{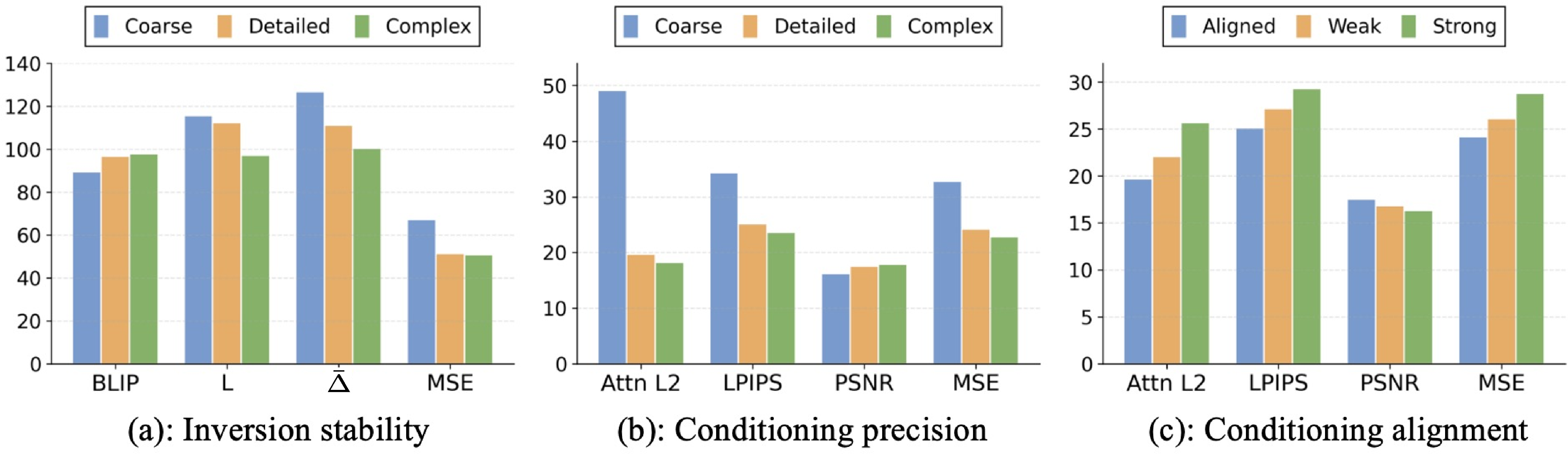}
    \caption{Textual conditioning stabilizes inversion dynamics and cross-branch attention coupling. (a) More precise textual conditioning better matches the source image, as indicated by higher BLIP similarity, and reduces directional deviation $\overline{\Delta}$, reconstruction MSE, and empirical Lipschitz constant $L$. (b) Increasing conditioning precision reduces cross-branch attention divergence, measured by Attn-L2, and improves background preservation metrics (LPIPS, PSNR, MSE). (c) Better source-target alignment shows the same trend, yielding more consistent attention coupling and faithful background preservation.}
    \label{fig:motivation}
\end{figure*}

\begin{figure}[t]
    \centering
    \includegraphics[width=1\linewidth]{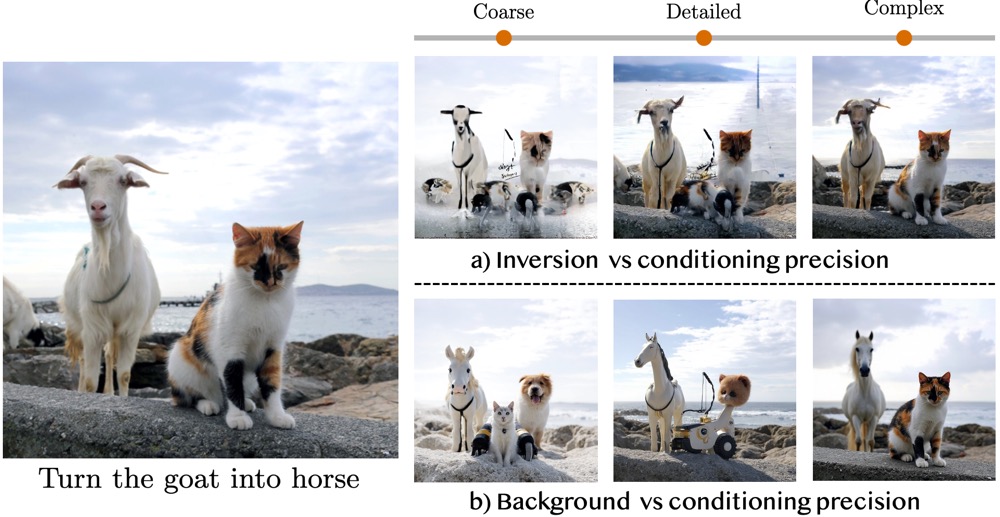}
    \caption{(a): Visualization of inversion reconstruction under different conditioning precision. (b): Editing results at varying prompt conditioning precision. As the conditioning precision increases, both the reconstruction accuracy and the preservation of background structure improve notably. }
    \label{fig:motivation_teaser}
\end{figure}

\subsection{Inversion Stability} \label{sec:inv_sta}
In inversion-based diffusion editing, the source image is mapped into the latent space by integrating the sampling ODE in the reverse direction with a discretized solver.
Inversion errors arise from forward–backward discretization discrepancies, which are amplified by local fluctuations in the velocity field. As illustrated in Fig.~\ref{fig:inversion_sta}, a stable velocity field is therefore crucial for accurate inversion. Prior analyses (e.g.,~\cite{chen2024trajectory,stanczuk2024diffusion}) suggest that the learned velocity field $\mathbf{v}_\theta$ approximates a local average direction toward the data manifold. Building on this geometric interpretation, we hypothesize that the precision of textual conditioning implicitly constrains the effective support of the conditional data manifold: imprecise conditioning leads to dispersed and less coherent velocity directions, whereas more precise and semantically grounded conditioning stabilizes the velocity field.

To empirically examine this hypothesis, we construct conditioning signals with progressively richer semantic descriptions by expanding the original prompts with additional image-grounded details, resulting in three levels of conditioning: \textbf{coarse}, \textbf{detailed}, and \textbf{complex}. At each timestep, we perturb the latent state $\mathbf{x}_{\sigma_t}$ with small Gaussian noise and measure the directional stability of the velocity field. 
Specifically, given the reference velocity $\mathbf{v}_{\sigma_t} = \mathbf{v}_\theta(\mathbf{x}_{\sigma_t},\sigma_t,c)$ along the inversion trajectory, we generate $N$ perturbed samples 
$\mathbf{x}_{\sigma_t}^{(i)} = \mathbf{x}_{\sigma_t} + \mathbf{\delta}^{(i)}$, where $\mathbf{\delta}^{(i)} \sim \mathcal{N}(0,\epsilon^2 I)$ and $i=1,\dots,N$, and compute the corresponding velocities $\mathbf{v}_{\sigma_t}^{(i)}$.
The cosine directional deviation is then defined as:
\[
\Delta = \frac{1}{T}\frac{1}{N}\sum_{t=1}^{T}\sum_{i=1}^{N}\left(1-\cos\left(\mathbf{v}_{\sigma_t}^{(i)},\mathbf{v}_{\sigma_t}\right)\right),
\]
where $\cos(\mathbf{a},\mathbf{b}) = \frac{\mathbf{a}^\top \mathbf{b}}{\|\mathbf{a}\|\,\|\mathbf{b}\|}$ denotes the cosine similarity between two vectors. This deviation measures the sensitivity of the velocity direction under local perturbations (see Appendix~B for details). A smaller $\Delta$ indicates smoother and more stable dynamics. We compute the average $\overline{\Delta}$ over an evaluation subset of PIE-Bench~\cite{ju2023direct}. As shown in Fig.~\ref{fig:motivation}(a), conditioning signals with higher semantic precision consistently yield smaller directional deviation $\overline{\Delta}$ and lower MSE reconstruction loss, which leads to more satisfied reconstruction results in Fig.~\ref{fig:motivation_teaser}(a).



Besides the empirical analysis of directional stability, we further provide a theoretical justification for the observed relationship between conditioning precision and inversion stability. Let $\mathbf{v}(\mathbf{x}, t; P)$ denote the velocity field conditioned on the textual signal $P$. We consider the pure inversion–reconstruction setting, where $\mathbf{x}_0$ is the input source image to be inverted, and $\mathbf{x}_{\mathrm{rec}}(0; P)$ is the image recovered by inverting $\mathbf{x}_0$ to the noise latent and integrating back to $t=0$ under the same conditioning $P$. Under mild smoothness assumptions, with both trajectories computed using the first-order explicit Euler solver with maximum step size $h$, the inversion–reconstruction error satisfies:
\[
\left\| \mathbf{x}_{\mathrm{rec}}(0; P) - \mathbf{x}_0 \right\|
\le 2\,\frac{C}{L}\left( e^{L T} - 1 \right)h,
\]
where $L = \sup_t \|J_v(\mathbf{x}, t)\|_2$ is the Lipschitz constant of the velocity field and
$C = \tfrac{1}{2}\sup_t \|J_v \mathbf{v} + \partial_t \mathbf{v}\|$ characterizes trajectory curvature (see Appendix~A for derivation). For first-order flow-based sampling with maximum step size $h$, a smaller Lipschitz constant $L$ leads to a tighter reconstruction bound.

We estimate the empirical Lipschitz constant via stochastic power iteration. As illustrated in Fig.~\ref{fig:motivation}(a), conditioning signals with higher semantic precision consistently yield smaller estimated $L$, thereby tightening the theoretical error bound and improving inversion stability.

\subsection{Effectiveness of Cross-Branch Attention Manipulation}

In the attention manipulation stage, we examine how textual conditioning influences cross-branch attention coupling during editing. Starting from the shared latent obtained from inversion, attention manipulation injects features from the source branch into the target branch. The construction of the conditioning signals influences how preserved structures and newly introduced semantics are integrated during this attention manipulation. 

\textbf{Conditioning precision and background preservation.}
Attention manipulation integrates structural information from the source branch with edit semantics from the target branch during generation. The precision of textual conditioning plays a critical role in this process: imprecise conditioning provides weak semantic guidance, which can lead to asymmetric attention features between the two branches and consequently inaccurate reconstruction of source image details. Following the experimental setting in Sec.~\ref{sec:inv_sta}, we analyze how conditioning precision affects cross-branch attention consistency and background preservation during editing. To quantify cross-branch attention divergence, we use Attn-L2, defined as the averaged L2 distance between corresponding source- and target-branch cross-attention maps, with the full computation provided in Appendix~C. As shown in Fig.~\ref{fig:motivation}(b), more precise textual conditioning reduces this attention divergence, thereby facilitating attention fusion and improving background preservation, as demonstrated in Fig.~\ref{fig:motivation_teaser}(b).

\textbf{Conditioning alignment.}
In addition to the precision, the structural alignment of textual conditioning between the source and target branches also influences cross-branch attention coupling. When the conditioning structures are misaligned—even if they describe semantically equivalent content—the corresponding attention features can become asymmetric during editing, leading to feature interference and degraded background preservation. To examine this effect, we construct prompt pairs with identical semantics but different levels of structural alignment. Starting from aligned source–target conditioning, we introduce \textbf{weak} and \textbf{strong} perturbations that progressively disrupt their alignment without changing their semantics. As shown in Fig.~\ref{fig:motivation}(c), better-aligned conditioning leads to lower cross-branch divergence between corresponding attention elements, resulting in improved background preservation during editing.

%% file: sec/4_method.tex
\section{Method}\label{sec:methods}

\begin{figure}[t]
    \centering
    \includegraphics[width=1\linewidth]{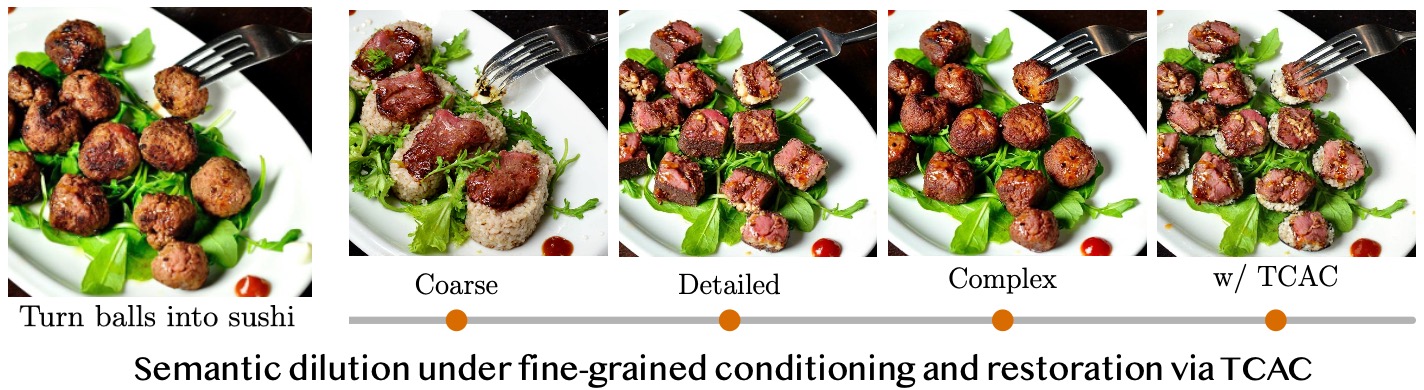}
    \caption{In certain cases, however, finer-grained textual conditioning may dilute the semantics of the core editing content—e.g., the sushi example shows that detailed and complex conditioning fail to clearly represent the intended sushi concept. Our proposed token-wise cross-branch attention control successfully restores the semantic fidelity of the edited content.}
    \label{fig:dilution}
\end{figure}

\begin{figure}[t]
    \centering
    \includegraphics[width=0.95\linewidth]{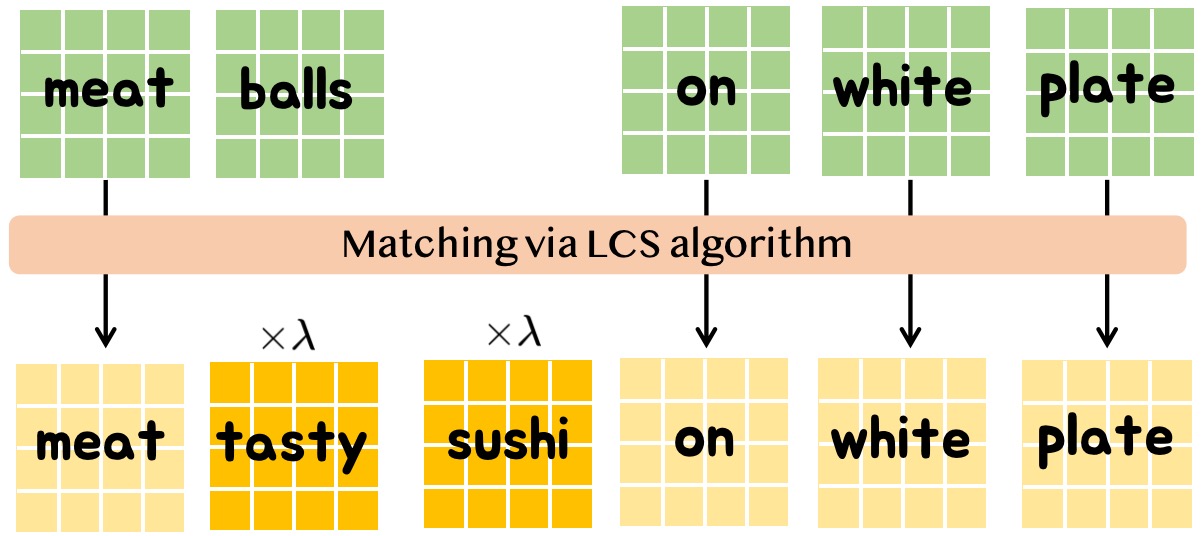}
    \caption{In our token-wise attention control, shared tokens are matched using the LCS algorithm. Subsequently, attention features for editing-driving and structure-preserving tokens are processed separately through \textit{reinforcement} and \textit{replacement}, respectively.}
    \label{fig:attention_control}
\end{figure}

Building upon the conditioning analysis in Sec.~4, we propose a conditioning-aware inversion-based editing framework shown that improves both inversion stability and cross-branch attention consistency. The framework contains two components: (1) \textit{conditioning refinement}, which increases conditioning precision while preserving structural alignment between source and target prompts; and (2) \textit{token-wise cross-branch attention control}, which separates edit-relevant and structure-preserving tokens and applies asymmetric attention modulation during attention manipulation. 

Refined conditioning improves inversion stability and background preservation by providing more precise and aligned textual guidance. However, it also increases token complexity in the prompt. Since attention weights are normalized across tokens, complex conditioning may dilute the contribution of edit-driving tokens, weakening the semantic strength of the intended edit. To address this issue, we introduce token-wise cross-branch attention control, which first separates edit-driving and structure-preserving tokens using a dynamic programming-based alignment algorithm, and then modulates their attention contributions asymmetrically during attention manipulation.

\begin{algorithm}[tb]
\caption{Token Alignment via LCS}
\label{alg:lcs-alignment}
\textbf{Input}: Target $\hat{P}_{tgt}$, Source $\hat{P}_{src}$, Tokenizer $\mathcal{T}$ \\
\textbf{Output}: Structure-preserving token indices $\overline{k}$ and edit-driving token indices $k$

\begin{algorithmic}[1]
\STATE $\texttt{target\_id} \gets \mathcal{T}.\texttt{encode}(\hat{P}_{tgt})$
\STATE $\texttt{src\_id} \gets \mathcal{T}.\texttt{encode}(\hat{P}_{src})$
\STATE $n \gets |\texttt{target\_id}|$, $m \gets |\texttt{src\_id}|$
\STATE Initialize $dp$ as zero matrix of size $(n{+}1) \times (m{+}1)$
\FOR{$i = 0$ to $n{-}1$}
  \FOR{$j = 0$ to $m{-}1$}
    \IF{$\texttt{target\_id}[i] = \texttt{src\_id}[j]$}
      \STATE $dp[i{+}1][j{+}1] \gets dp[i][j] + 1$
    \ELSE
      \STATE $dp[i{+}1][j{+}1] \gets \max(dp[i][j{+}1], dp[i{+}1][j])$
    \ENDIF
  \ENDFOR
\ENDFOR
\STATE Backtrack from $dp[n][m]$ to obtain $(\mathcal{K}_{\mathrm{tgt}},\mathcal{K}_{\mathrm{src}})$
\STATE $\overline{k} \gets \mathcal{K}_{\text{tgt}}$ \hfill \COMMENT{structure-preserving tokens}
\STATE $k \gets \{0,\ldots,n{-}1\} \setminus \mathcal{K}_{\text{tgt}}$ \hfill \COMMENT{edit-driving tokens}
\STATE \textbf{return} $\overline{k}, k$
\end{algorithmic}
\end{algorithm}

\subsection{Conditioning Refinement (CR)}

As shown in Sec.\ref{sec:motivation}, conditioning precision and structural alignment influence inversion stability and cross-branch attention coupling. To improve both properties, we refine the conditioning signals by expanding prompts with additional contextual attributes, structured descriptions and background information while preserving the shared semantic structure between the source and target prompts.
In practice, conditioning refinement can be instantiated using external language models to generate more descriptive prompts. However, our framework is not tied to any specific refinement model, and alternative strategies that improve conditioning precision while maintaining source–target alignment can also be adopted.

\subsection{Token-wise Cross-Branch Attention Control (TCAC)}

While refined conditioning improves structural consistency, it also increases token complexity and may dilute edit semantics under attention normalization (as shown in Fig.~\ref{fig:dilution}). To mitigate this effect, we partition conditioning tokens into two groups using the Longest Common Subsequence (LCS) algorithm in Alg.~\ref{alg:lcs-alignment} between the tokenized sequences of refined source and target prompts: \textit{structure-preserving tokens} $\overline{k}$ shared across the two prompts, and \textit{edit-driving tokens} $k$ that appear only in the target prompt. Given this token partition, we apply asymmetric cross-branch attention control as demonstrated in Fig.~\ref{fig:attention_control}. During the cross-branch editing process, structure-preserving tokens follow standard feature injection from the source branch into the target branch to preserve background consistency, while edit-driving tokens $k$ are reinforced by a amplification factor $\lambda$ during attention computation to compensate for normalization-induced dilution:
\[
A[:, k_i] \leftarrow \lambda A[:, k_i].
\]
This asymmetric modulation preserves background structures while maintaining strong editing semantics under dense conditioning.

%% file: sec/5_experiments.tex
\section{Experiments}

\begin{table}[t]
  \centering
  \resizebox{\columnwidth}{!}{
  \begin{tabular}{lcccc}
    \toprule
    \textbf{Method} & PSNR ↑ & LPIPS ↓ & MSE ↓ & SSIM ↑\\
    \midrule
    Euler & 20.94 & 167.49 & 151.25 & 81.02\\
    Euler + \textbf{CR} 
      & 22.19 \textcolor{red}{\scriptsize{(+5.98\%)}} 
      & 139.5 \textcolor{red}{\scriptsize{(-16.71\%)}} 
      & 112.46 \textcolor{red}{\scriptsize{(-25.61\%)}} 
      & 83.18 \textcolor{red}{\scriptsize{(+2.67\%)}} \\
    \hline
    RF-Solver & 22.77 & 146.82 & 119.86 & 79.79\\
    RF-Solver + \textbf{CR} 
      & 24.52 \textcolor{red}{\scriptsize{(+7.68\%)}} 
      & 109.11 \textcolor{red}{\scriptsize{(-25.69\%)}} 
      & 67.87 \textcolor{red}{\scriptsize{(-43.37\%)}} 
      & 83.8 \textcolor{red}{\scriptsize{(+5.04\%)}} \\
    \hline
    FireFlow & 16.62 & 268.8 & 354.74 & 65.88\\
    FireFlow + \textbf{CR} 
      & 18.16 \textcolor{red}{\scriptsize{(+9.25\%)}} 
      & 220.76 \textcolor{red}{\scriptsize{(-17.88\%)}} 
      & 231.9 \textcolor{red}{\scriptsize{(-34.61\%)}} 
      & 70.46 \textcolor{red}{\scriptsize{(+6.94\%)}} \\
    \hline
    FP Inversion & 20.81 & 147.11 & 177.25 & 79.98\\
    FP Inversion + \textbf{CR}
      & 22.96 \textcolor{red}{\scriptsize{(+10.33\%)}} 
      & 96.9 \textcolor{red}{\scriptsize{(-34.14\%)}} 
      & 124.04 \textcolor{red}{\scriptsize{(-29.98\%)}} 
      & 84.47 \textcolor{red}{\scriptsize{(+5.62\%)}} \\
    \bottomrule
  \end{tabular}}
  \vspace{2pt}
  \caption{Quantitative comparison for the effectiveness of conditioning refinement (\textbf{CR}) in inversion and reconstruction. }
  \label{tab:inversion}
\end{table}

\begin{figure*}[t]
    \centering
    \includegraphics[width=\linewidth]{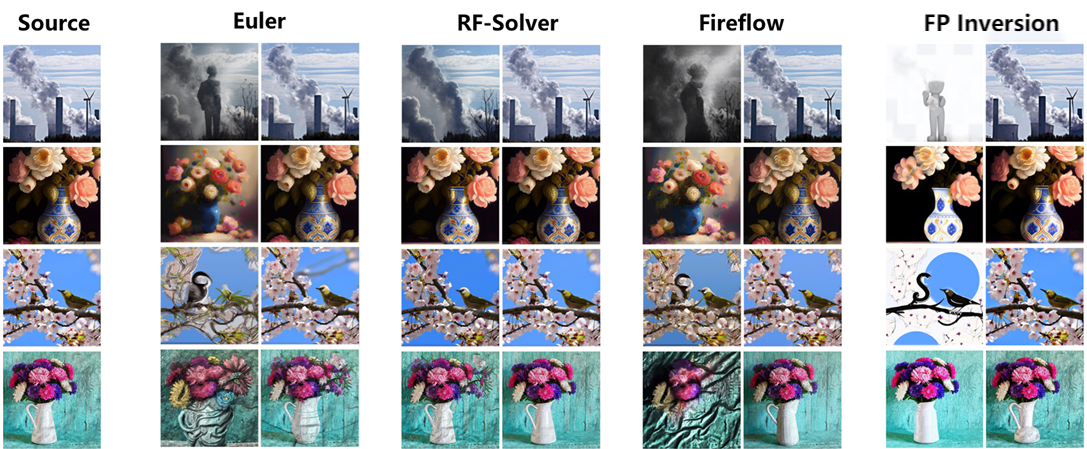}
    \caption{Inversion and reconstruction results using different solvers, including Euler, RF-Solver \cite{wang2025taming}, FireFlow \cite{deng2025fireflow}, and FP Inversion \cite{meiri2023fixed}. For each method, reconstruction is performed with the coarse conditioning (left) and the refined conditioning (right).}
    \label{fig:inversion}
\end{figure*}

\begin{table*}[!ht]
\centering
\resizebox{\textwidth}{!}{%
\begin{tabular}{lclccccccccccc}
\toprule
\multirow{2}{*}{\textbf{Method}} & \multirow{2}{*}{\textbf{Mask}} & \multirow{2}{*}{\textbf{Model}} & \multicolumn{1}{c}{\textbf{Structure}} & \multicolumn{4}{c}{\textbf{Background Preservation}} & \multicolumn{2}{c}{\textbf{CLIP Similarity}} & \multicolumn{2}{c}{\textbf{CLIPSim*}} \\
\cmidrule(lr){4-4} \cmidrule(lr){5-8} \cmidrule(lr){9-10} \cmidrule(lr){11-12}
& & & Distance $\downarrow$ & PSNR $\uparrow$ & LPIPS $\downarrow$ & MSE $\downarrow$ & SSIM $\uparrow$ & Whole $\uparrow$ & Edited $\uparrow$ &  Whole $\uparrow$ & Edited $\uparrow$ \\
\midrule
Uni-edit (ICLR'26) \cite{jiao2025uniedit} & Y & FLUX & 10.35 & 28.97 & 63.35 & 20.98 & 90.67 & 25.71 & 22.24 & 27.35 & 22.71 \\
Uni-edit+\textbf{CR} & Y & FLUX
  & 8.83
  & 30.17
  & 53.29
  & 16.76
  & 91.66
  & 25.65
  & 22.18
  & 27.76
  & 22.90 \\
KV-edit (CVPR'25) \cite{zhu2025kvedit} & Y & FLUX & 1.70 & 34.85 & 12.4 & 5.42 & 95.70 & 24.28 & 20.14 & 26.30 & 20.42 \\
KV-edit+\textbf{CR} & Y & FLUX 
  & 1.72
  & 34.82
  & 12.5
  & 5.43
  & 95.71
  & 24.75
  & 20.62
  & 26.98
  & 21.22\\
\midrule
P2P \cite{hertz2023p2p} & N & SD1.4 & 71.34 & 17.85 & 209.1 & 220.8 & 71.44 & 25.13 & 22.47 & 25.07	& 22.14 \\
PnP \cite{tumanyan2023plug} & N & SD1.4 & 27.43 & 22.26 & 113.58 & 84.13 & 79.24 & 25.45 & 22.54 & 25.65 & 22.79 \\
MasaCtrl \cite{cao2023masactrl} & N & SD1.4 & 28.38 & 22.17 & 106.62 & 86.97& 79.67 & 23.96	& 22.16 & 25.75 & 21.85 \\
P2P-Zero \cite{parmar2023zero} & N & SD1.4 & 61.68 & 20.44 & 172.22 & 144.12 & 74.67 & 22.8 & 20.54 & - & - \\
Rf-Inversion (IClR'25) \cite{rout2025semantic} & N & FLUX & 20.84 & 22.56 & 171.5 & 85.77 & 74.98 & 25.3 & 22.14 & 26.64 & 22.54 \\
Rf-Edit (ICML'25)\cite{wang2025taming} & N & FLUX & \textbf{18.59} & 23.39 & 137.2 & 70.09 & 81.26 & 25.97 & 22.8 & 27.09 & 23.27 \\
FireFlow (ICML'25) \cite{deng2025fireflow} & N & FLUX & 28.3 & 23.08 & 128.79 & 71.07 & 81.15 & 26.16 & 22.85 & 27.47 & 23.42 \\
\textbf{\ourName} & N & FLUX & 23.20 & \textbf{24.03} & \textbf{98.60} & \textbf{59.61} & \textbf{85.75} & \textbf{26.18} & \textbf{23.07} & \textbf{28.38} & \textbf{23.90} \\
\bottomrule
\end{tabular}}
\vspace{2pt}
\caption{Quantitative comparison on PIE-Bench.}
\label{tab:quantitative_comparison}
\end{table*}

\begin{table*}[!ht]
\centering
\resizebox{\textwidth}{!}{%
\begin{tabular}{llcccccccccc}
\toprule
\multirow{2}{*}{\textbf{Method}} & \multirow{2}{*}{\textbf{Model}} & \multicolumn{1}{c}{\textbf{Structure}} & \multicolumn{4}{c}{\textbf{Background Preservation}} & \multicolumn{2}{c}{\textbf{CLIP Similarity}} & \multicolumn{2}{c}{\textbf{CLIP Similarity*}}  \\
\cmidrule(lr){3-3} \cmidrule(lr){4-7} \cmidrule(lr){8-9} \cmidrule(lr){10-11}
& & Distance $\downarrow$ & PSNR $\uparrow$ & LPIPS $\downarrow$ & MSE $\downarrow$ & SSIM $\uparrow$ & Whole $\uparrow$ & Edited $\uparrow$ &  Whole $\uparrow$ & Edited $\uparrow$  \\
\midrule
Base & FLUX & 24.13 & 23.99 & 108.79 & 62.48 & 85.01 & 25.91 & 22.78 & 27.18 & 23.16 \\
+\textbf{CR} & FLUX &  23.28 & 24.09 & 97.57 & 60.15 & 85.61 & 25.92 & 22.83 & 28.24 & 23.78  \\
+\textbf{CR} +\textbf{TCAC}  & FLUX & 23.20 & 24.03 & 98.60 & 59.61 & 85.75 & 26.18 & 23.07 & 28.38 & 23.90  \\
\bottomrule
\end{tabular}
}
\vspace{2pt}
\caption{Ablation study on PIE-Bench.}
\label{tab:ablation}
\end{table*}

\subsection{Setup}

\paragraph{Baselines and Implementations.}
We compare \ourName with a wide range of attention manipulation-based editing methods across two major categories: (1) early methods \cite{hertz2023p2p,tumanyan2023plug,cao2023masactrl,li2023zero} built on U-Net-based T2I models \cite{rombach2022ldm}, and (2) recent methods \cite{rout2025semantic,wang2025taming,deng2025fireflow} built on DiT-based T2I models~\cite{esser2024scaling,black-forest2024flux}. 
All baselines use official implementations and default settings. To directly demonstrate the effect of our method, we conduct primary evaluations without external masks. In practice, our conditioning-control framework is orthogonal to mask supervision and can be readily integrated into existing mask-guided editing pipelines~\cite{zhu2025kvedit,jiao2025uniedit}, further improving their performance (Tab.~\ref{tab:quantitative_comparison}). Unless otherwise specified, we adopt FLUX.1-dev as the base model with its standard 25 sampling steps for both inversion and editing. 
In attention manipulation-based editing, the manipulation interval has a strong influence on the editing outcome. Longer manipulation intervals inject more structural information from the source branch, improving background preservation but weakening semantic fidelity. To balance structural preservation and semantic modification, we apply cross-branch attention manipulation within the timestep interval $[0,4]$.  For conditioning refinement, we use \texttt{Gemini-2.5-Pro} to expand the original prompts with additional image-grounded descriptions. More details are provided in Appendix~D.

\paragraph{Evaluation Datasets and Metrics.}
Following our baselines, we evaluate on PIE-Bench~\cite{ju2023direct}, which contains 700 editing examples across 10 categories. Each example contains source/target prompts, an editing instruction, and a binary mask specifying the edited region, allowing separate evaluation of edited and preserved areas. Following prior work, we measure performance from three perspectives:
(1) structural consistency (DINO feature distance),
(2) background preservation (PSNR, SSIM, LPIPS, MSE outside edited region),
and (3) semantic alignment (CLIP similarity).Since conditioning signals in PIE-Bench are often minimal and foreground-focused, raw CLIP similarity may under-reflect global semantic coherence. To obtain a more comprehensive assessment of conditioning alignment, we additionally report CLIPSim*, computed using descriptive reference conditioning that captures the full semantic content of the edited image. The reference conditioning used for evaluation is included in the released code.

\subsection{Comparisons}

\paragraph{Inversion and Reconstruction.}
To verify the geometric analysis in Sec.~4, we evaluate inversion and reconstruction under refined conditioning signals across multiple inversion solvers performed on the same base model~\cite{black-forest2024flux}. As shown in Tab.~\ref{tab:inversion}, conditioning refinement consistently reduces reconstruction errors (PSNR, SSIM, MSE, LPIPS) across different solvers. This supports our theoretical claim that conditioning precision stabilizes the velocity field, effectively reducing the accumulation of inversion errors. Qualitative results in Fig.~\ref{fig:inversion} further illustrate that coarse conditioning leads to trajectory deviation and reconstruction artifacts, while refined conditioning yields smoother dynamics and improved reconstruction fidelity.

\paragraph{Image Editing.}
We evaluate editing performance under mask-free settings to directly demonstrate the effect of our methods. As shown in Tab.~\ref{tab:quantitative_comparison}, our framework achieves superior performance across both structural preservation and semantic alignment metrics compared to existing attention manipulation baselines in a mask-free setting. Moreover, when integrated into mask-guided editing pipelines, our conditioning refinement schema further improves editing performance, demonstrating its compatibility and complementarity with external structural supervision. Notably, background preservation metrics and semantic similarity inherently form a trade-off.
Methods favoring reconstruction tend to under-edit, while aggressive semantic conditioning often sacrifices structure. Our approach achieves a balanced regime by combining conditioning refinement with token-wise cross-Branch attention control. Qualitative comparisons in Fig.~\ref{fig:editing} further demonstrate improved structural coherence and semantic precision.

\subsection{Ablation Study}

We analyze the contribution of each component in our \ourName editing framework. Starting from a base attention manipulation-based editing setup, we progressively introduce (i) \textbf{C}onditioning \textbf{R}efinement(\textbf{CR}) and (ii) \textbf{T}oken-wise \textbf{C}ross-branch \textbf{A}ttention \textbf{C}ontrol(\textbf{TCAC}). As shown in Tab.~\ref{tab:ablation}, conditioning refinement brings substantial improvements in background preservation (e.g., LPIPS reduces by 10\%), consistent with its role in stabilizing inversion geometry and enabling more accurate recovery of source image details.
The subsequent introduction of token-wise cross-branch attention control tackles the normalization-induced dilution caused by conditioning refinement. As illustrated in Fig.~\ref{fig:dilution}, refined textual conditioning with dense tokens can dilute core editing semantics (e.g., the sushi example), and TCAC successfully recovers the semantic strength of edit-driving tokens, leading to improved CLIP similarity while maintaining the background preservation quality. More qualitative results are provided in the Appendix~D.

\subsection{Additional Analysis}


\subsubsection{Impact of Conditioning Refinement}

\begin{table*}[!ht]
\centering
\resizebox{\textwidth}{!}{%
\begin{tabular}{lccccccccccc}
\toprule
\multirow{2}{*}{\textbf{Method}} & \multicolumn{1}{c}{\textbf{Structure}} & \multicolumn{4}{c}{\textbf{Background Preservation}} & \multicolumn{2}{c}{\textbf{CLIP Similarity}} & \multicolumn{2}{c}{\textbf{CLIP Similarity*}}  \\
\cmidrule(lr){2-2} \cmidrule(lr){3-6} \cmidrule(lr){7-8} \cmidrule(lr){9-10}
& Distance $\downarrow$ & PSNR $\uparrow$ & LPIPS $\downarrow$ & MSE $\downarrow$ & SSIM $\uparrow$ & Whole $\uparrow$ & Edited $\uparrow$ &  Whole $\uparrow$ & Edited $\uparrow$  \\
\midrule
P2P & 38.36 & 20.89 & 138.71 & 129.94 & 81.02 & 26.49 & 23.20 & 27.15 & 23.21 \\
P2P+CR & 38.07 & 21.32 & 127.23 & 116.18 & 82.26 & 26.36 & 23.12 & 28.48 & 23.79  \\
Masa & 23.32 & 24.18 & 106.87 & 60.23 & 85.29 & 25.83 & 22.71 & 27.11 & 23.13  \\
Masa+CR & 22.93 & 24.17 & 97.71 & 58.98 & 85.67 & 25.87 & 22.79 & 28.19 & 23.76  \\
\bottomrule
\end{tabular}
}
\vspace{2pt}
\caption{Impact of our conditioning refinement on different editing pipeline.}
\label{tab:cr}
\end{table*}

To demonstrate the general applicability of conditioning refinement across different attention manipulation strategies, we analyze its effect on several representative inversion-based editing pipelines. We apply the same conditioning refinement strategy to several representative inversion-based editing frameworks. 
For a fair comparison under the same backbone, we re-implement the classical P2P (QK replacement) and MasaCtrl (KV replacement) attention manipulation strategies on the FLUX model. As shown in Tab.~\ref{tab:cr}, conditioning refinement consistently improves editing performance across different pipelines. These results suggest that conditioning refinement acts as a general improvement to conditioning quality and can benefit multiple inversion-based editing frameworks.

\subsubsection{Conditioning Refinements with Different Vision-Language Models}

\begin{table*}[!ht]
\centering
\resizebox{\textwidth}{!}{%
\begin{tabular}{lcccccccccc}
\toprule
\multirow{2}{*}{\textbf{Model}} & \multicolumn{1}{c}{\textbf{Structure}} & \multicolumn{4}{c}{\textbf{Background Preservation}} & \multicolumn{2}{c}{\textbf{CLIP Similarity}} & \multicolumn{2}{c}{\textbf{CLIP Similarity*}}  \\
\cmidrule(lr){2-2} \cmidrule(lr){3-6} \cmidrule(lr){7-8} \cmidrule(lr){9-10}
& Distance $\downarrow$ & PSNR $\uparrow$ & LPIPS $\downarrow$ & MSE $\downarrow$ & SSIM $\uparrow$ & Whole $\uparrow$ & Edited $\uparrow$ & Whole $\uparrow$ & Edited $\uparrow$  \\
\midrule
No Refine & 24.13 & 23.99 & 108.79 & 62.48 & 85.01 & 25.91 & 22.78 & 27.18 & 23.16 \\
GPT5 & 23.94 & 23.95 & 103.54 & 60.77 & 85.13 & 26.17 & 23.10 & 28.23 & 23.91 \\
Gemini2.5-Pro & 23.28 & 24.09 & 97.57 & 60.15 & 85.61 & 25.92 & 22.83 & 28.24 & 23.78 \\
\bottomrule
\end{tabular}
}
\vspace{2pt}
\caption{Effectiveness of conditioning refinement with GPT5 and Gemini2.5-Pro.}
\label{tab:model}
\end{table*}

To examine whether conditioning refinement depends on a specific vision-language model, we generate refined prompts using different VLMs for the PIE-Bench dataset and evaluate the resulting editing performance under our pipeline. As shown in Tab.~\ref{tab:model}, prompts refined by different VLMs consistently improve editing performance compared to the original prompts. Although the absolute performance varies slightly across models, the overall trend remains consistent across evaluation metrics. These results suggest that conditioning refinement is model-agnostic and can generally improve conditioning quality for inversion-based image editing.

\subsubsection{Latency Analysis.}
The average runtimes for the standard inversion and editing stages on PIE-Bench are 31.78\,s and 7.34\,s, respectively. Our proposed conditioning refinement requires 11.97\,s, resulting in a total runtime of 51.09\,s per sample. This indicates that the conditioning refinement step accounts for only a small portion of the overall pipeline and is not the main latency bottleneck.

%% file: sec/6_conclusion.tex
\section{Conclusion}

In this paper, we present \ourName, a conditioning-aware framework for inversion-based diffusion image editing.
By analyzing textual conditioning as a geometric constraint on the diffusion velocity field, we show that the formulation of textual conditioning influences inversion stability and cross-branch attention consistency. Building upon this insight, our framework refines conditioning signals to stabilize diffusion dynamics and introduces token-wise cross-branch attention control to balance structural preservation and semantic modification. Together, these components enable improved inversion fidelity and editing accuracy without retraining the underlying generative model. Although \ourName achieves strong and consistent performance, the timestep interval for attention manipulation still requires manual selection. This limitation is shared by most existing inversion-based editing methods that rely on attention manipulation. An important direction for future work is to develop adaptive scheduling strategies that automatically adjust the timestep interval for attention manipulation based on input-dependent conditioning characteristics.

\paragraph{Supplementary Material.}
Additional details, complete proofs, and extended experimental results are provided in the supplementary material available at \url{https://github.com/zju-pi/SimEdit}.

%% file: appendix.tex
\section*{Appendix}

\subsection*{A. Reconstruction Bound} \label{appendix:recon-bound}

In this appendix, we provide the complete derivation of the inversion–reconstruction error bound stated in the main paper. We begin by introducing global smoothness assumptions on the learned velocity field and then derive the global error bound for Euler discretization of the corresponding ODE flow. We then describe our practical estimator for the local Lipschitz constant based on a memory-efficient stochastic power-iteration scheme using Jacobian–vector (JVP) and vector–Jacobian (VJP) products.

\subsubsection*{Preliminaries}

We consider the latent trajectory governed by the flow
\begin{equation}
    \dot{\mathbf{x}}(t)
    = \mathbf{v}(\mathbf{x}(t), t; P),
    \qquad t \in [0,T],
\end{equation}
where $\mathbf{v}$ is the velocity field conditioned on the textual prompt
$P$.  
Both inversion (integrating $t\!:\!0\!\rightarrow\!T$) and reconstruction
($T\!\rightarrow\!0$) employ the \emph{same} velocity field and the same discretization scheme. Although flow-based generative models typically employ non-uniform time schedules, 
each update can be viewed as a first-order Euler integration step with a locally defined step size $h$.

\paragraph{Assumption A.1 (Global mild smoothness).}
The velocity field $\mathbf{v}(\mathbf{x},t;P_{\mathrm{src}})$ is continuously
differentiable in both $\mathbf{x}$ and $t$, and its Jacobian and total time
derivative are uniformly bounded over a domain
$\Omega\subset\mathbb{R}^d$ that contains all trajectories of interest.
That is, there exist constants $L,C>0$ such that for all
$\mathbf{x}\in\Omega$ and $t\in[0,T]$,
\begin{align}
    & \|J_v(\mathbf{x}, t)\|_2 \le L,
    \label{eq:lip-appendix}
    \\
    & \|J_v(\mathbf{x}, t)\mathbf{v}(\mathbf{x}, t)
      + \partial_t \mathbf{v}(\mathbf{x}, t)\|
    \le 2C.
    \label{eq:curvature-appendix}
\end{align}
Equivalently,
\begin{equation}
\begin{aligned}
    L 
    &= \sup_{\mathbf{x}\in\Omega, t\in[0,T]}
        \|J_v(\mathbf{x}, t)\|_2,
    \\
    C 
    &= \frac12 \sup_{\mathbf{x}\in\Omega, t\in[0,T]}
       \| J_v(\mathbf{x}, t)\mathbf{v}(\mathbf{x}, t)
         + \partial_t\mathbf{v}(\mathbf{x}, t)\|.
\end{aligned}
\end{equation}
Here, $L$ is a global Lipschitz constant of the velocity field, while $C$ controls the
magnitude of the second-order term
$\frac{\mathrm{d}}{\mathrm{d}t}\mathbf{v}
   = J_v\mathbf{v} + \partial_t\mathbf{v}$,
which determines the curvature of Euler trajectories.

\subsubsection*{Inversion--Reconstruction Error Bound}

Under the smoothness assumptions stated above, the classical global error of explicit Euler integration with step size $h$ satisfies:
\begin{equation}
    \|\mathbf{x}_k - \mathbf{x}(t_k)\|
    \le
    \frac{C}{L}\bigl(e^{L t_k}-1\bigr)h.
    \label{eq:euler-single-pass}
\end{equation}

Since both inversion ($0\to T$) and reconstruction ($T\to 0$) integrate the
same velocity field using Euler discretization, the total error can be bounded
by the sum of the two Euler integration errors.

\begin{proposition}[Inversion--Reconstruction Error]
\label{prop:recon-bound}
Under the bounds in Eqs.~\eqref{eq:lip-appendix}--\eqref{eq:curvature-appendix}, the
reconstruction obtained by integrating forward to $T$ and back satisfies:
\begin{equation}
\boxed{
\left\|
\mathbf{x}_{\mathrm{rec}}(0; P)
- 
\mathbf{x}_0
\right\|
\le
2\,\frac{C}{L}\left(e^{LT}-1\right)\ h_{max}.
}
\label{eq:recon-bound-appendix}
\end{equation}
\end{proposition}
Here, $h_{\max}=\max_k h_k$ denotes the maximum step size along the discretized trajectory. For flow models with $T=1$, the exponential dependence on $L$ dominates the bound; therefore smaller Lipschitz constants directly lead to tighter reconstruction guarantees.

\subsubsection*{Estimating the Lipschitz Constant $L$}

Direct computation of 
\[
L = sup_{\mathbf{x},t} ||J_v(\mathbf{x},t)||_2
\]
is infeasible because the Jacobian $J_v\in\mathbb{R}^{d\times d}$ for diffusion
models has dimension factor $d$ in the tens of thousands.
Forming or storing this matrix is impossible, and computing its spectral norm exactly would require a prohibitively expensive singular value decomposition.

To obtain a tractable approximation, we estimate $\|J_v\|_2$ using
\emph{stochastic power iteration}.
The spectral norm satisfies
\[
\|J_v\|_2 = \max_{\|u\|=1} \|J_v u\|,
\]
and repeated application of $J_v^\top J_v$ drives any vector toward the dominant singular vector.  
Crucially, this requires only Jacobian--vector products (JVPs) and
vector--Jacobian products (VJPs), both of which can be computed through
automatic differentiation without forming $J_v$ explicitly. This yields an tractable and memory-efficient estimator as demonstrated in Alg.~\ref{alg:jv}, which is applicable to high-dimensional velocity fields.

\paragraph{Trajectory-level estimation.}
We estimate local Jacobian spectral norms along the inversion trajectory.
For every \texttt{stride} steps, we run $M$ iterations of power iteration and
obtain a local singular value estimate $\hat{\lambda}_k$.
To ensure robustness while avoiding pathological outliers caused by rare numerical spikes, we report the
empirical Lipschitz constant as the $95$th percentile:
\[
L_{\mathrm{emp}}
=
\mathrm{P95}\bigl(\{\hat{\lambda}_k\}\bigr).
\]
In all experiments, we use \texttt{stride}\,$=2$ and $M=3$, which provides a
practical balance between estimation accuracy and computational overhead.

\vspace{4pt}
\begin{algorithm}[h!]
\caption{Stochastic Spectral-Norm Estimation of $\|J_v\|_2$ via JVP/VJP}
\label{alg:power-iteration}
\begin{algorithmic}[1]
\REQUIRE Inversion trajectory $\{(\mathbf{x}_k,t_k)\}$, stride $s$, iterations $M$
\STATE $\mathcal{K} \gets \{\, k \mid k \equiv 0 \pmod{s} \,\}$
\FOR{each $k \in \mathcal{K}$}
    \STATE Initialize $u_0 \sim \mathcal{N}(0,I)$ and normalize
    \FOR{$m = 0$ \textbf{to} $M-1$}
        \STATE $v_m \gets J_v(\mathbf{x}_k,t_k)\,u_m$ \hfill \textit{(JVP)}
        \STATE $w_m \gets J_v(\mathbf{x}_k,t_k)^\top v_m$ \hfill \textit{(VJP)}
        \STATE $u_{m+1} \gets w_m / \|w_m\|$
    \ENDFOR
    \STATE $\hat{\lambda}_k \gets \|J_v(\mathbf{x}_k,t_k)\,u_M\|$
\ENDFOR
\RETURN $L_{\mathrm{emp}} = \mathrm{P95}\left(\{\hat{\lambda}_k : k\in\mathcal{K}\}\right)$
\end{algorithmic}
\label{alg:jv}
\end{algorithm}

\section*{B. Inversion Stability}

This experiment examines how conditioning precision affects inversion stability. 
In this subsection, we describe the experimental procedure used to compute the directional deviation $\Delta$, which measures the sensitivity of the velocity field to local perturbations along the inversion trajectory.

To empirically examine this effect, we construct conditioning signals with progressively richer semantic descriptions by expanding the original prompts with additional image-grounded details. 
This procedure produces three levels of textual conditioning signals: 
$P_1$ (coarse), $P_2$ (detailed), and $P_3$ (comprehensive). 
For each conditioning signal $P$, we perform the following steps:

\begin{enumerate}
    \item \textbf{Inversion Trajectory Construction:} 
    Given the prompt $P$ and its corresponding source image $\mathbf{x}_0$, we run the forward process in flow matching conditioned on $P$ to obtain the inversion trajectory $\{\mathbf{x}_{\sigma_t}^{\text{inv}}\}_{t=0}^{T}$ with the velocity field $\{\mathbf{v}_{\sigma_t}(\mathbf{x}_{\sigma_t}^{\text{inv}},\sigma_t,P)\}_{t=0}^{T}$ along this trajectory.
    
    \item \textbf{Perturbation:} 
    At each timestep $t$, we generate $N$ perturbed samples $\{\mathbf{x}_{\sigma_t}^{(i)}\}_{i=1}^{N}$ by adding Gaussian noise:
    \[
    \mathbf{x}_{\sigma_t}^{(i)} = \mathbf{x}_{\sigma_t}^{\text{inv}} + \delta^{(i)}, \quad \delta^{(i)} \sim \mathcal{N}(0, \epsilon^2 \mathbf{I})
    \]
    In this experiment, we set $\epsilon=0.005$. Similar conclusions can also be drawn under different $\epsilon$ settings.

    \item \textbf{Velocity Prediction:} 
    For each perturbed latent $\mathbf{x}_{\sigma_t}^{(i)}$, we compute its velocity $\mathbf{v}$ conditioned on the signal $P$:
    \[
    \mathbf{v}_{\sigma_t}^{(i)} = \mathbf{v}_\theta(\mathbf{x}_{\sigma_t}^{(i)}, \sigma_t, P)
    \]
    
    \item \textbf{Angular Deviation Calculation:}  
    To assess the stability of the velocity field, we measure the mean angular deviation $\Delta$ between the velocities $\mathbf{v}_{\sigma_t}^{(i)}$ on perturbed latents $\mathbf{x}_{\sigma_t}^{(i)}$ and the reference velocity $\mathbf{v}_{\sigma_t}$ along the inversion trajectory:
    \[
        \Delta = \frac{1}{T}\frac{1}{N}\sum_{t=1}^{T}\sum_{i=1}^{N}\left(1-\cos\left(\mathbf{v}_{\sigma_t}^{(i)},\mathbf{v}_{\sigma_t}\right)\right),
    \]
    A smaller $\Delta$ indicates lower velocity deviation under perturbations and thus reflects a more stable and consistent velocity field.

    \item \textbf{Statistical Comparison:} 
    For computational efficiency, we conduct this analysis on an official evaluation subset of \textbf{PIE-Bench}. For each image, we compute $\Delta$ and report the dataset-level average $\overline{\Delta}$.
\end{enumerate}

\section*{C. Effectiveness of Cross-Branch Attention Manipulation} \label{appendix:attn-consistency}
In this experiment, we study how conditioning precision and alignment affect the distance between attention maps of the source and target branches during attention manipulation.
For each editing instruction, we construct conditioning signals with different levels of structural alignment between the source and target branches. 
Starting from aligned source–target conditioning pairs, we introduce controlled perturbations to the conditioning structure while preserving the underlying semantic content. 
This allows us to analyze how conditioning alignment affects attention behavior during editing.

To isolate effects purely attributable to attention behavior—rather than inversion—we directly sample the latent $\mathbf{x}_T$ from a Gaussian distribution as the initialization for source/target branches, instead of performing inversion process.

\begin{figure*}[t]
    \centering
    \includegraphics[width=1\linewidth]{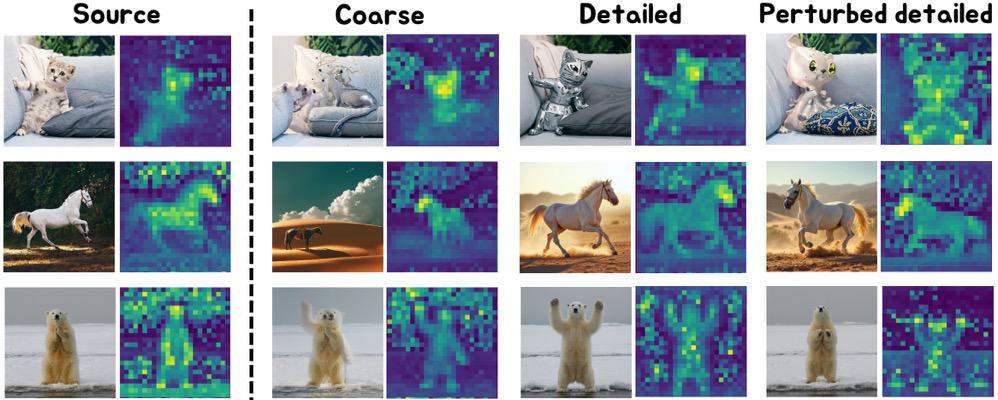}
    \caption{This figure presents the cross-attention maps of core content—specifically, the words "cat", "horse", and "bear" shown in the three respective rows—in the source branch and the target branch performed on different settings of conditioning signals, along with the corresponding editing results. The figure demonstrates that more precise and aligned textual conditioning lead to more consistent cross-attention maps and more accurate editing outcomes.}
    \label{fig:attn_consistency2}
\end{figure*}

\paragraph{Attention distance.}
To measure the consistency between attention features in the source and target branches, we compute the distance between their corresponding attention maps. Here, we provide a detailed explanation of how we compute the distance between attention maps.

\noindent In the FLUX model, the attention map is computed as
\[
A = \mathrm{softmax}(QK^\top / \sqrt{d}),
\]
where the query and key matrices are constructed by concatenating prompt and image features,
\[
Q = \begin{bmatrix} Q_p, Q_{\text{img}} \end{bmatrix},
\quad
K = \begin{bmatrix} K_p, K_{\text{img}} \end{bmatrix}.
\]
This yields a block-structured attention map:
\[
A = 
\begin{bmatrix}
Q_pK_p^\top & Q_pK_{\text{img}}^\top \\
Q_{\text{img}}K_p^\top & Q_{\text{img}}K_{\text{img}}^\top
\end{bmatrix},
\]
where the top-left and bottom-right blocks correspond to self-attention among text tokens and image patches, respectively, and the off-diagonal blocks encode cross-attention between them.
We focus on the top-right cross-attention block
\[
A^{\text{cross}} = Q_pK_{\text{img}}^\top,
\]
which captures how each text token attends to image patches. In our experiment, we quantify attention consistency at the cross-attention block. Concretely, we adopt the following analysis pipeline to measure cross-attention consistency between the source and target branches:

\begin{enumerate}
\item \textbf{Cross-attention map extraction:}
For each layer $l$, timestep $t$, and attention head $h$, we extract the full cross-attention block $A^{\text{cross}}_{l,t,h}$ for both the source and target branches.

\item \textbf{Aggregation across layers, timesteps, and heads:}
We average the extracted cross-attention maps over all layers, timesteps, and heads to obtain aggregated cross-attention maps
\[
\mathbf{A}_{\text{src}}^{\text{cross}}
\quad\text{and}\quad
\mathbf{A}_{\text{tgt}}^{\text{cross}}
\]
for the source and target branches, respectively.

\item \textbf{Attention dissimilarity metric:}
We compute the (optionally normalized) $\ell_2$ distance between the aggregated source and target cross-attention maps:
\[
D_{\text{attn}} = \left\| \mathbf{A}_{\text{src}}^{\text{cross}} - \mathbf{A}_{\text{tgt}}^{\text{cross}} \right\|_2,
\]
where the maps are flattened into vectors before computing the norm.

\item \textbf{Statistical comparison:}
We evaluate $D_{\text{attn}}$ for each example in the dataset and report the average within each experimental group (e.g., different conditioning precision or syntactic variants) for comparison.
\end{enumerate}

In practice, higher attention consistency facilitates more stable and semantically coherent injection of attention features during editing, leading to improved editing results as shown in Fig.~\ref{fig:attn_consistency2}. 
These observations suggest that more precise and better-aligned conditioning signals help improve the effectiveness of attention manipulation in diffusion-based image editing.

\section*{D. Implementation Details}

Our implementation is based on FLUX.1-dev, with both inversion and editing performed using the default 25 sampling steps. For evaluation on PIE-Bench, attention manipulation is applied at timesteps $[0, 4]$ to balance background preservation and editing fidelity. To ensure reproducibility, we fix the random seed to $2$ for all experiments involving stochasticity.  All experiments are conducted on two NVIDIA A800 GPUs under CUDA~12.0. The base FLUX.1-dev model requires about $48$GB of GPU memory for processing $512\times512$ images in PIE-Bench. We use Ubuntu~18.04.5 as the operating system, with relevant software libraries Python~3.10, PyTorch~2.5.1+cu121 and diffusers~0.31.0.

\subsection*{Conditioning Refinement}

In our implementation, refined conditioning signals are generated using a vision-language model based on the original source–target prompt pairs provided in PIE-Bench. 
Given the source image and the original prompts, the model produces more detailed textual descriptions of the source image and refined target prompts that expand the editing instruction while remaining semantically consistent with the source content. 
The refinement step aims to complement parts of the source image that are not explicitly described in the original prompts. In particular, the refined conditioning signals provide additional descriptions of background context, secondary objects and attributes, as well as scene composition and style information that may be omitted in the original conditioning. To improve inversion stability and attention consistency, the generated source and target conditioning signals follow a shared structural format, ensuring that their semantic components remain aligned. 
This produces more descriptive and better-aligned conditioning signals for inversion-based editing.

\begin{figure*}
    \centering
    \includegraphics[width=1\linewidth]{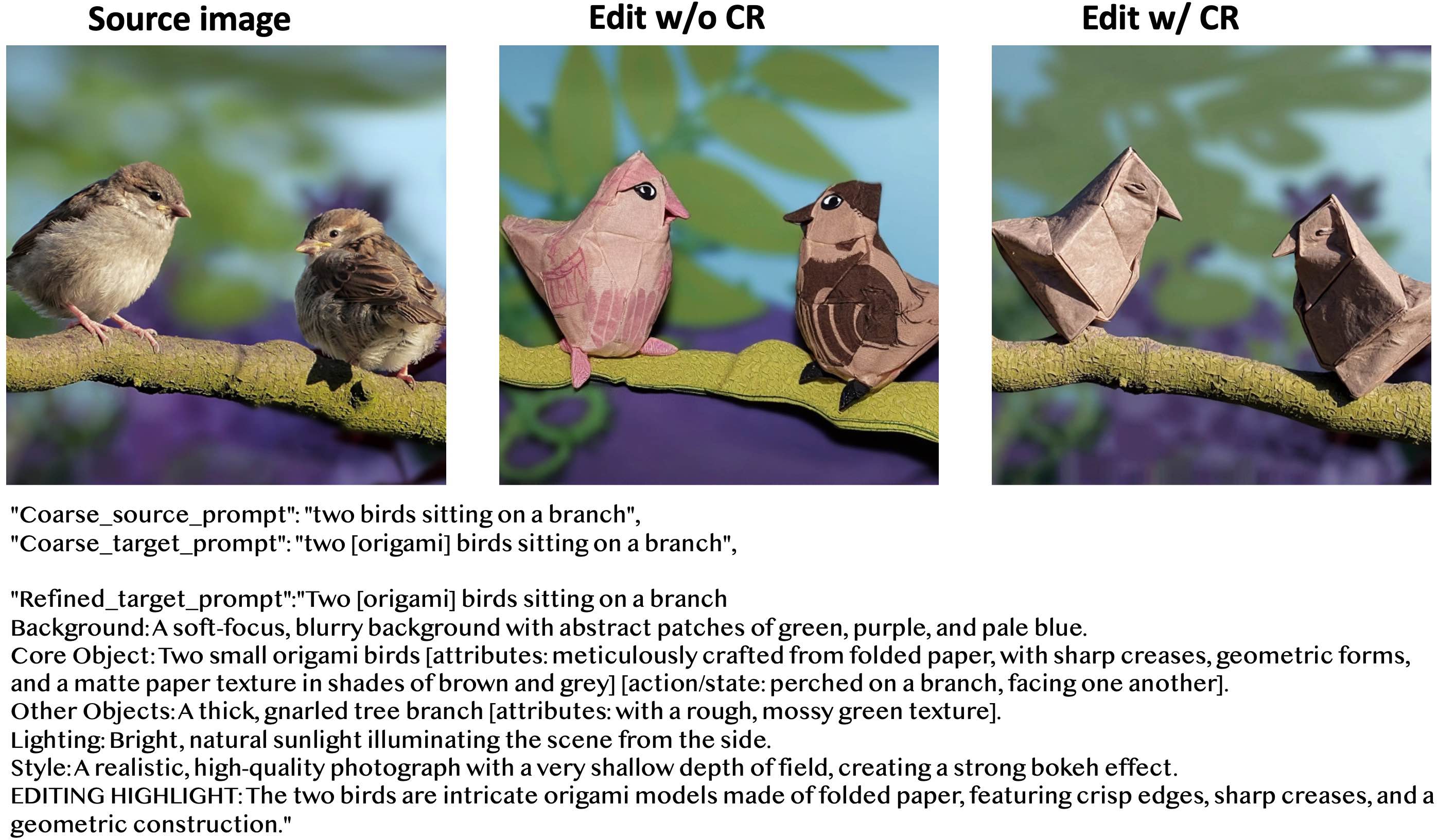}
    \caption{Example 1 of the effectiveness of our conditioning refinement strategy in image editing.}
    \label{fig:pe1}
\end{figure*}

\begin{figure*}
    \centering
    \includegraphics[width=1\linewidth]{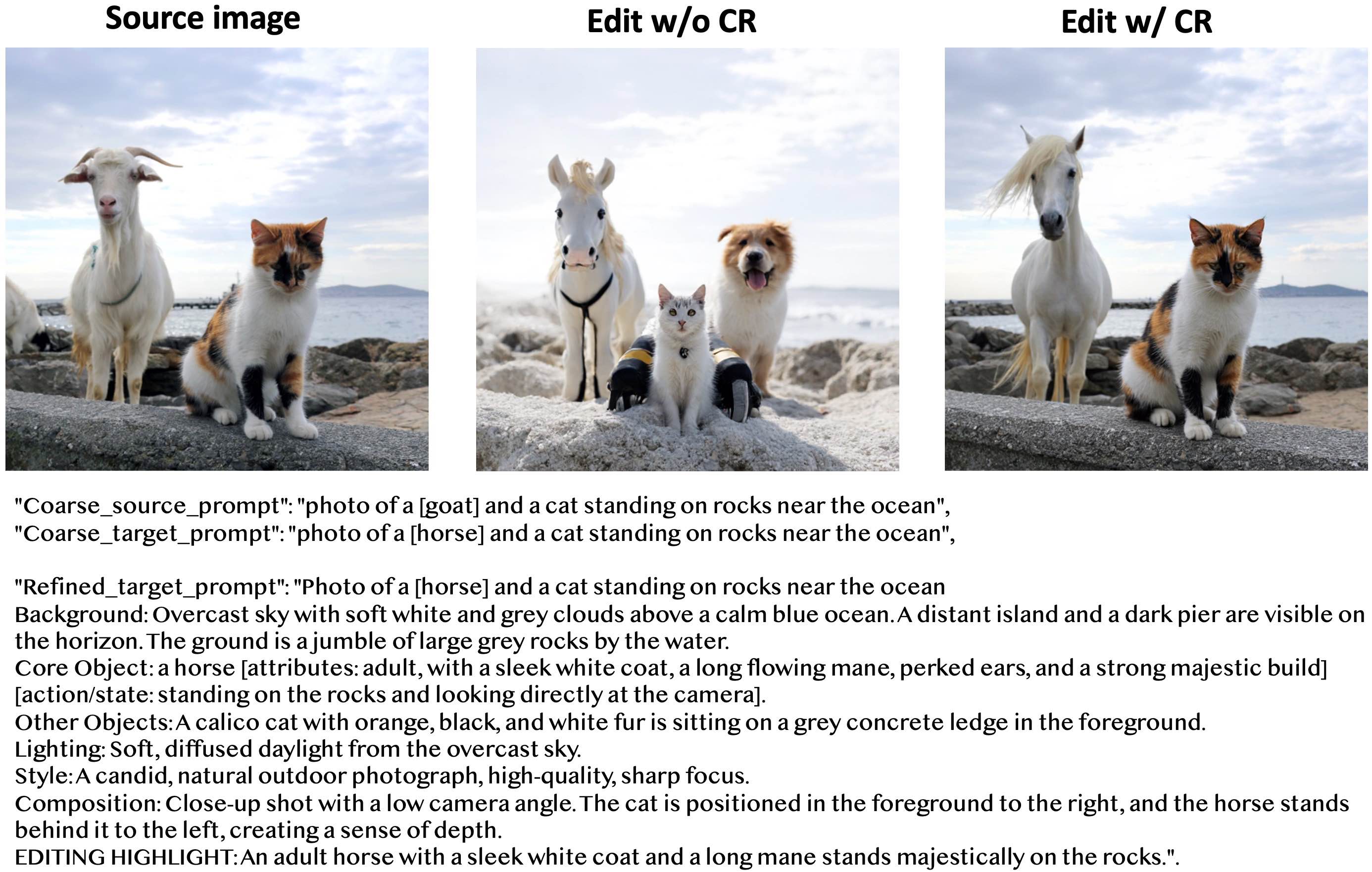}
    \caption{Example 2 of the effectiveness of our conditioning refinement strategy in image editing.}
    \label{fig:pe2}
\end{figure*}

\begin{figure*}
    \centering
    \includegraphics[width=1\linewidth]{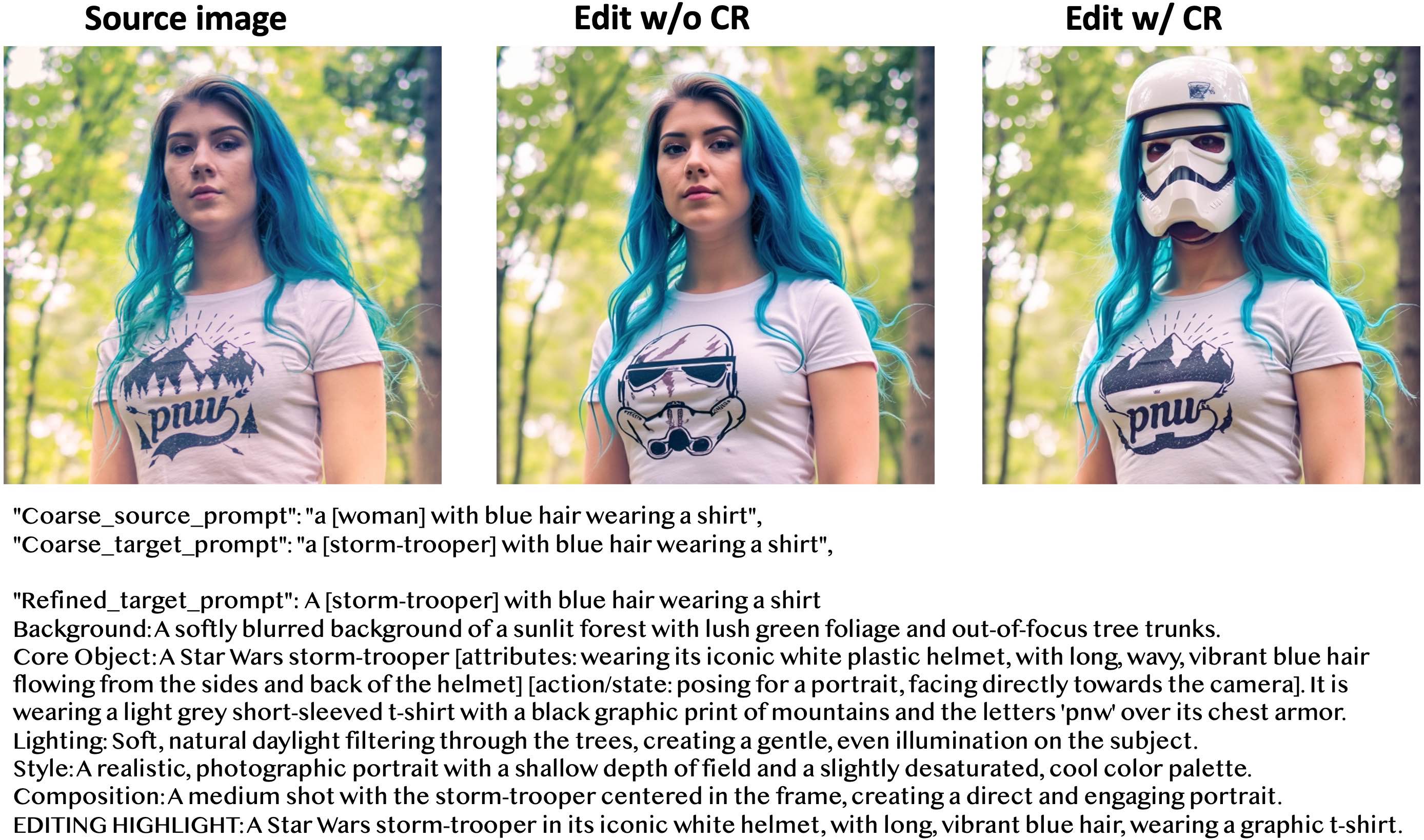}
    \caption{Example 3 of the effectiveness of our conditioning refinement strategy in image editing.}
    \label{fig:pe3}
\end{figure*}

\begin{figure*}
    \centering
    \includegraphics[width=1\linewidth]{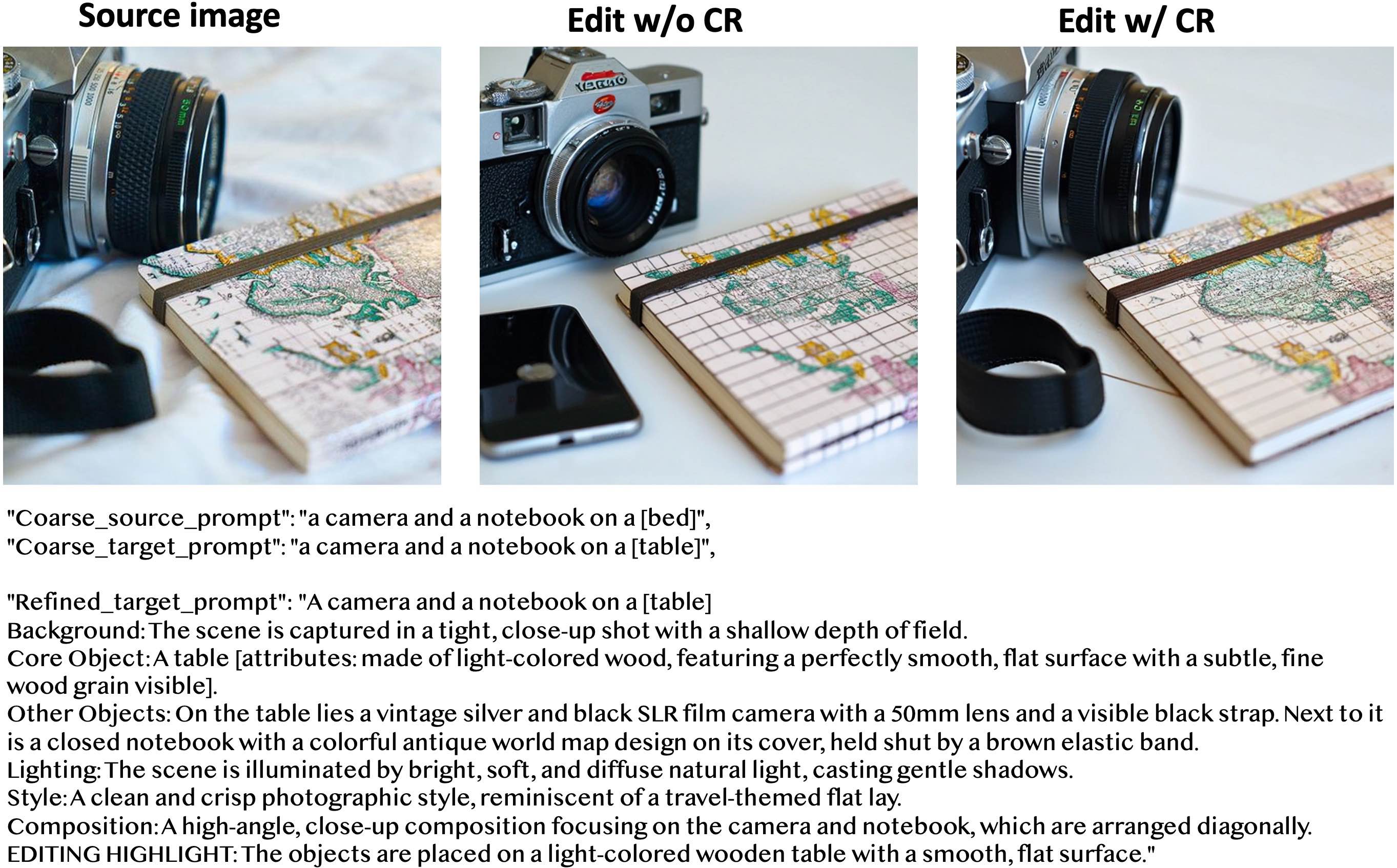}
    \caption{Example 4 of the effectiveness of our conditioning refinement strategy in image editing.}
    \label{fig:pe4}
\end{figure*}

\begin{figure*}
    \centering
    \includegraphics[width=1\linewidth]{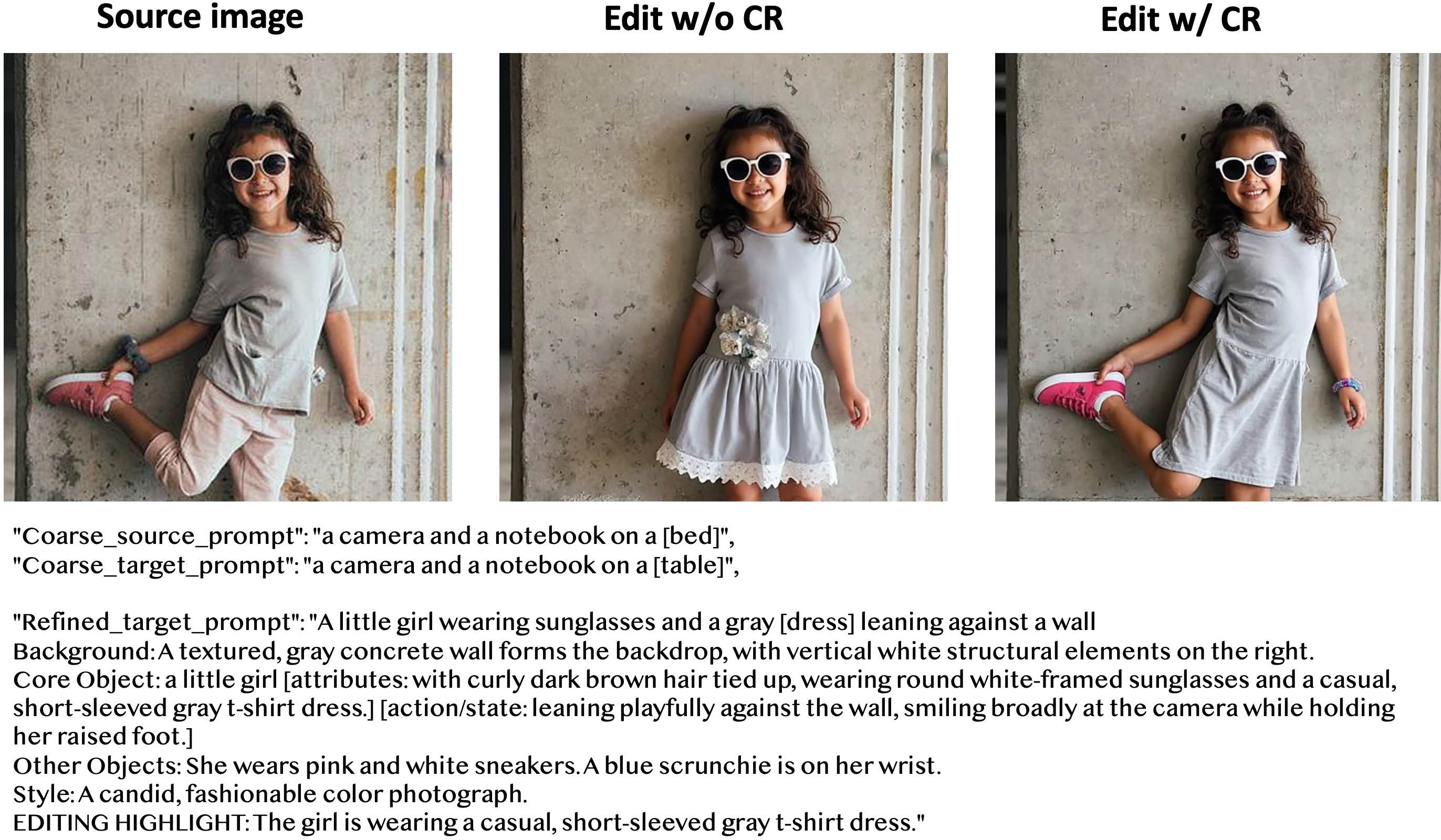}
    \caption{Example 5 of the effectiveness of our conditioning refinement strategy in image editing.}
    \label{fig:pe5}
\end{figure*}

\begin{figure*}
    \centering
    \includegraphics[width=1\linewidth]{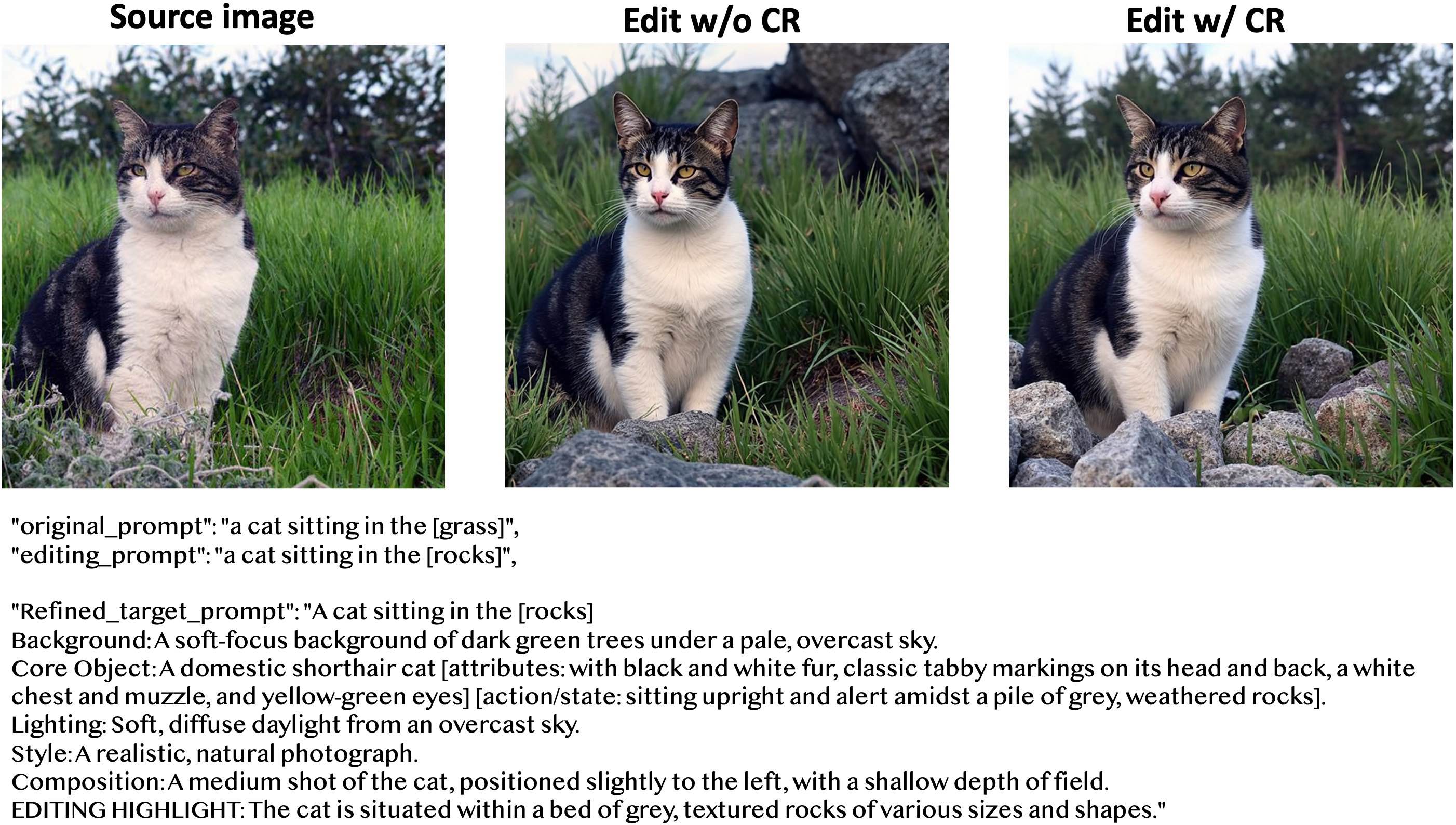}
    \caption{Example 6 of the effectiveness of our conditioning refinement strategy in image editing.}
    \label{fig:pe6}
\end{figure*}

For reproducibility, the system prompts used for conditioning generation are included in the released codebase. 
We also provide the refined prompts for the full PIE-Bench dataset generated using \texttt{Gemini-2.5-Pro} with the same system prompts. Fig.~\ref{fig:pe1}–\ref{fig:pe6} present examples of original and refined conditioning signals together with the resulting editing outputs. We observe that more descriptive conditioning signals generally improve background preservation and structural consistency in the edited images.

\begin{figure*}[t]
    \centering
    \includegraphics[width=1\linewidth]{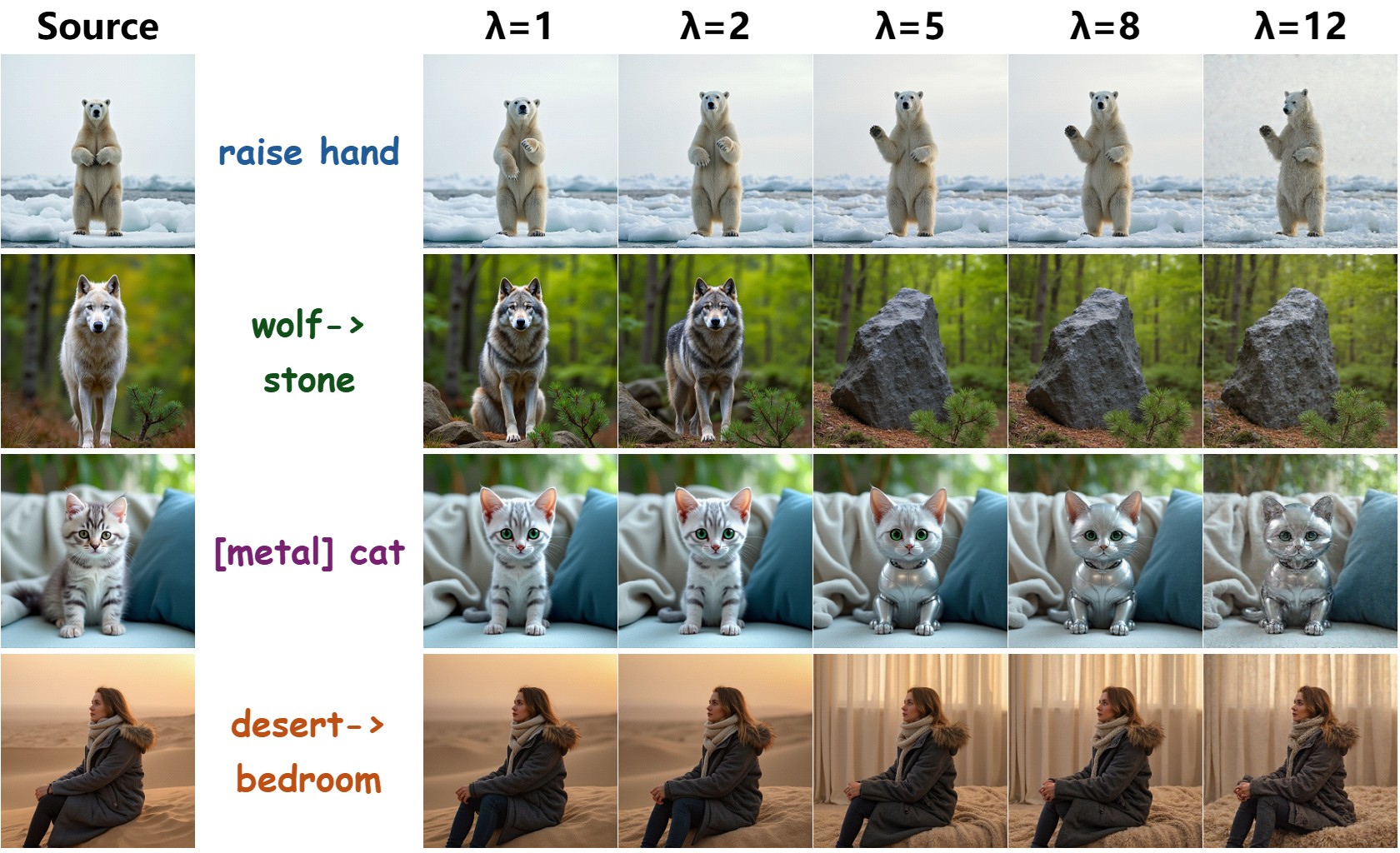}
    \caption{Qualitative results of editing under different scales of attention reinforcement. A properly chosen scale $\lambda$ effectively guides the generation of the desired content in the edited output, while an excessively large $\lambda$ compromises the model’s generative capacity, leading to degraded results.}
    \label{fig:ae}
\end{figure*}

\subsection*{Time Interval for Attention Manipulation}
The timestep interval over which attention manipulation is applied has a significant impact on editing behavior. To study this effect, we evaluate several manipulation intervals and report the editing performance under different manipulation intervals in Tab.~\ref{tab:interval}. Experiment results demonstrate that as the manipulation interval becomes longer, more structural information from the source branch is injected into the generation process. 
This leads to improved background preservation, as reflected by better PSNR, SSIM, LPIPS, and MSE scores. However, the stronger structural injection also reduces semantic alignment with the target prompt, as indicated by the decrease in CLIP similarity. This reveals a trade-off between background preservation and editing fidelity. 
Based on the results in Tab.~\ref{tab:interval}, we adopt the interval $[0-4]$, which provides a balanced compromise between structural preservation and semantic editing.

\begin{table}[t]
\centering
\resizebox{\linewidth}{!}{%
\begin{tabular}{ccccccccccc}
\toprule
\multirow{2}{*}{\textbf{Interval}} & \multicolumn{1}{c}{\textbf{Structure}} & \multicolumn{4}{c}{\textbf{Background Preservation}} & \multicolumn{2}{c}{\textbf{CLIP Similarity}} & \multicolumn{2}{c}{\textbf{CLIP Similarity*}} \\
\cmidrule(lr){2-2} \cmidrule(lr){3-6} \cmidrule(lr){7-8} \cmidrule(lr){9-10}
& Distance $\downarrow$ & PSNR $\uparrow$ & LPIPS $\downarrow$ & MSE $\downarrow$ & SSIM $\uparrow$ & Whole $\uparrow$ & Edited $\uparrow$ & Whole $\uparrow$ & Edited $\uparrow$ \\
\midrule
0--2 & 37.73 & 21.27 & 132.95 & 108.86 & 81.92 & 26.42 & 23.45 & 28.54 & 24.23 \\
0--4 & 23.20 & 24.03 & 98.60 & 59.61 & 85.75 & 26.18 & 23.07 & 28.38 & 23.90 \\
0--6 & 15.86 & 26.16 & 78.18 & 38.83 & 88.15 & 25.48 & 22.38 & 27.86 & 23.41 \\
0--8 & 11.23 & 27.85 & 63.30 & 27.20 & 89.96 & 24.92 & 21.75 & 27.42 & 22.88 \\
\bottomrule
\end{tabular}
}
\vspace{2pt}
\caption{Effect of manipulation interval on editing performance on PIE-Bench.}
\label{tab:interval}
\end{table}

\subsection*{Token-level attention control}
\subsubsection*{Token partition}
In order to perform accurate token-level attention control, we employ a Longest Common Subsequence (LCS) algorithm to align tokens in source and target prompts, which identifies the edit-relevant tokens $k$ and edit-irrelevant tokens $\overline{k}$. In this section, we provide the complete separating-token algorithm, including backtracking, in Alg.~\ref{alg:lcs-alignment-full}. The aligned token indices $\mathcal{K}_{\text{tgt}}$ identify the subset of target prompt tokens that are semantically shared with the source, which we designate as edit-irrelevant tokens $\overline{k}$ for attention replacement. Conversely, the remaining tokens in the target prompt that are not part of the LCS, which are identified as edit-relevant tokens $k$ and serve as the focus of attention reinforcement. 

To examine the stability of our LCS-based token matching, we analyze the shared-token coverage between the source and target prompts under LCS alignment. Tab.~\ref{tab:alignment_stats} reports the resulting statistics across the dataset. The average coverage reaches 90.2\% for the source prompts and 78.3\% for the target prompts, with the P10 coverage remaining above 77.2\% and 65.4\%, respectively. The slightly lower target coverage is expected, as the refined target prompts contain additional tokens describing and emphasizing newly introduced semantics. This behavior arises because conditioning refinement produces semantically aligned source–target prompt pairs, allowing LCS matching to reliably capture the shared semantic components between them. These results indicate that LCS provides stable token-level alignment on the refined prompts and can robustly separate shared and edit-relevant tokens.

\begin{table}[t]
\centering
\begin{tabular}{lcccc}
\toprule
Prompt & Mean & Median & P10 & P90 \\
\midrule
Source & 90.2\% & 92.8\% & 77.2\% & 99.3\% \\
Target & 78.3\% & 78.9\% & 65.4\% & 90.6\% \\
\bottomrule
\end{tabular}
\vspace{2pt}
\caption{Statistics of shared-token coverage rates.}
\label{tab:alignment_stats}
\end{table}

\begin{algorithm}[tb]
\caption{Token Alignment via LCS}
\label{alg:lcs-alignment-full}
\textbf{Input}: Target $\hat{P}_{tgt}$, Source $\hat{P}_{src}$, Tokenizer $\mathcal{T}$ \\
\textbf{Output}: Structure-preserving token indices $\overline{k}$ and edit-driving token indices $k$

\begin{algorithmic}[1]
\STATE $\texttt{target\_id} \gets \mathcal{T}.\texttt{encode}(\hat{P}_{tgt})$
\STATE $\texttt{src\_id} \gets \mathcal{T}.\texttt{encode}(\hat{P}_{src})$
\STATE $n \gets |\texttt{target\_id}|$, $m \gets |\texttt{src\_id}|$
\STATE Initialize $dp$ as zero matrix of size $(n{+}1) \times (m{+}1)$
\FOR{$i = 0$ to $n{-}1$}
  \FOR{$j = 0$ to $m{-}1$}
    \IF{$\texttt{target\_id}[i] = \texttt{src\_id}[j]$}
      \STATE $dp[i{+}1][j{+}1] \gets dp[i][j] + 1$
    \ELSE
      \STATE $dp[i{+}1][j{+}1] \gets \max(dp[i][j{+}1], dp[i{+}1][j])$
    \ENDIF
  \ENDFOR
\ENDFOR
\STATE $\mathcal{K}_{\text{tgt}} \gets [\,]$, $\mathcal{K}_{\text{src}} \gets [\,]$
\STATE $i \gets n$, $j \gets m$
\WHILE{$i > 0$ and $j > 0$}
  \IF{$\texttt{target\_id}[i{-}1] = \texttt{src\_id}[j{-}1]$}
    \STATE Prepend $i{-}1$ to $\mathcal{K}_{\text{tgt}}$
    \STATE Prepend $j{-}1$ to $\mathcal{K}_{\text{src}}$
    \STATE $i \gets i{-}1$, $j \gets j{-}1$
  \ELSIF{$dp[i{-}1][j] \ge dp[i][j{-}1]$}
    \STATE $i \gets i{-}1$
  \ELSE
    \STATE $j \gets j{-}1$
  \ENDIF
\ENDWHILE
\STATE $\overline{k} \gets \mathcal{K}_{\text{tgt}}$ \hfill \COMMENT{structure-preserving tokens}
\STATE $k \gets \{0,\ldots,n{-}1\} \setminus \mathcal{K}_{\text{tgt}}$ \hfill \COMMENT{edit-driving tokens}
\STATE \textbf{return} $\overline{k}, k$
\end{algorithmic}
\end{algorithm}

\begin{figure*}
    \centering
    \includegraphics[width=1\linewidth]{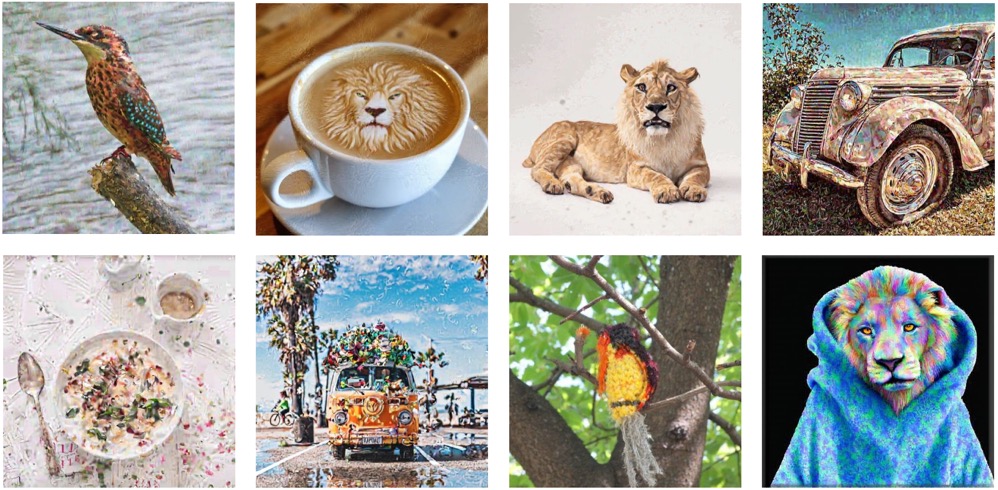}
    \caption{The degraded editing results of inversion-based editing on the PIE-Bench dataset when setting an emphasize scale of $\lambda = 5$ for attention reinforcement.}
    \label{fig:degredation}
\end{figure*}

\subsubsection*{Attention Reinforcement}

Fig.~\ref{fig:ae} illustrates the effect of attention reinforcement in the proposed token-wise cross-branch attention control. As shown in Fig.~\ref{fig:ae}, when refined conditioning signals become more descriptive, the increased token complexity may dilute the influence of edit-driving tokens during attention computation. As a result, the generated output may fail to incorporate the intended semantic modification ($\lambda=1$). In practice, attention reinforcement increases the relative influence of edit-driving tokens in the attention computation, allowing the desired editing semantics to be more effectively injected into the generation process. As $\lambda$ increases, the effect of attention reinforcement becomes progressively stronger, and the desired content gradually appears in the edited output, demonstrating the effectiveness of the proposed mechanism. However, when $\lambda$ becomes too large, we observe noticeable degradation in image quality. We attribute this behavior to excessive attention scaling, which causes the attention maps during generation to exceed the stable operating range of the base model and thereby compromises the generative capability of the FLUX backbone. This effect is further amplified in inversion-based editing. Because the sampling process starts from inverted latents rather than Gaussian noise, the accumulated inversion errors make the subsequent generation trajectory more sensitive to strong attention perturbations. Consequently, excessive attention reinforcement is more likely to cause visual artifacts under this setting (as illustrated in Fig.~\ref{fig:degredation}).  Based on these observations, we adopt a relatively moderate reinforcement scale of $\lambda = 3$ in all experiments, which effectively enhances editing semantics while avoiding visual artifacts or image degradation.

\section*{E. More qualitative results}

